\documentclass[sn-mathphys]{sn-jnl}

\usepackage{comment}
\usepackage{epstopdf}

\usepackage{subfigure}
\usepackage{indentfirst}
\usepackage{xcolor}
\usepackage[normalem]{ulem}

\usepackage{textcomp,booktabs}
\usepackage{pifont}

\jyear{2023}%

\theoremstyle{thmstyleone}%
\theoremstyle{thmstyletwo}%

\theoremstyle{thmstylethree}%

\begin{document}

\title{FEMOT: Multi-Object Tracking using Frame and Event Cameras}

\author[1]{\fnm{Shiao} \sur{Wang}}\email{e24101001@stu.ahu.edu.cn} 
\author[1]{\fnm{Xiao} \sur{Wang*}}\email{xiaowang@ahu.edu.cn}
\author[1]{\fnm{Chao} \sur{Wang}}\email{w853023886@126.com}
\author[1]{\fnm{Yitao} \sur{Li}}\email{2414333816@qq.com}
\author[1]{\fnm{Menghao} \sur{Liu}}\email{3528252469@qq.com}
\author[1]{\fnm{Bo} \sur{Jiang}}\email{jiangbo@ahu.edu.cn}
\author[2,5]{\fnm{Yaowei} \sur{Wang}}\email{wangyw@pcl.ac.cn}
\author[2,3,4]{\fnm{Yonghong} \sur{Tian}}\email{yhtian@pku.edu.cn}
\author[1]{\fnm{Jin} \sur{Tang}}\email{tangjin@ahu.edu.cn}

\affil[1]{\orgdiv{School of Computer Science and Technology, Anhui University, Hefei, 230601, China}}
\affil[2]{\orgdiv{Peng Cheng Laboratory, Shenzhen, China}} 
\affil[3]{\orgdiv{National Key Laboratory for Multimedia Information Processing, School of Computer Science, Peking University, China}}
\affil[4]{\orgdiv{School of Electronic and Computer Engineering, Shenzhen Graduate School, Peking University, China}}
\affil[5]{\orgdiv{Harbin Institute of Technology, Shenzhen, China}}

\abstract{
Conventional RGB cameras have been widely used in multi-object tracking due to their ability to capture rich appearance and semantic information. However, their performance is often degraded under complex real-world challenges, such as motion blur, low illumination, and overexposure. Bio-inspired event cameras offer high temporal resolution and high dynamic range, providing complementary cues under extreme scenarios. Nevertheless, RGB-event multi-object tracking remains underexplored due to the lack of large-scale and well-annotated datasets.
To address this issue, we propose FEMOT, a large-scale RGB-event multi-object tracking dataset that covers diverse real-world scenarios and 14 challenging attributes. With both RGB and event data as well as high-quality annotations, FEMOT provides a reliable platform for systematically evaluating RGB-event multi-object tracking methods. Based on FEMOT, we retrain and evaluate over ten strong trackers, thereby establishing a comprehensive benchmark for future research. Furthermore, we propose FEMOTR, a multimodal tracking framework that decouples RGB and event features and fuses them in the frequency domain, thereby effectively exploiting their complementary characteristics for robust object localization and identity association. Extensive experiments on FEMOT and DSEC-MOT datasets demonstrate the effectiveness of the proposed method.
The source code and benchmark dataset have been released on~\url{https://github.com/Event-AHU/FEMOT}. 
}


\maketitle

\section{Introduction}

Multi-Object Tracking (MOT) is a fundamental task in computer vision and artificial intelligence, aiming to localize multiple objects and maintain their identities over time. It plays a crucial role in many real-world applications, such as autonomous driving, intelligent transportation, robotics, and video surveillance. With the rapid advances in deep neural networks, frame-based MOT methods~\cite{zhang2021fairmot, zhang2022bytetrack, gao2023memotr, gao2025multiple, ren2024samba, cao2025topic} have made remarkable progress on recent large-scale benchmark datasets~\cite{milan2016mot16, dendorfer2020mot20, sun2022dancetrack, cui2023sportsmot, dave2020tao, yu2020bdd100k}. Equipped with powerful object detectors, discriminative appearance embeddings, and effective data association strategies, existing trackers have demonstrated reliable performance in many well-conditioned scenarios. For example, MeMOTR~\cite{gao2023memotr} exploits memory-enhanced trajectory representations to address long-term association, whereas MOTIP~\cite{gao2025multiple} reformulates association from a matching problem into an ID prediction task, both contributing to improved performance of RGB-based trackers.

However, achieving robust MOT in complex real-world environments remains a challenging problem, e.g., low illumination, backlighting, target occlusion, dense crowds, camera motion, adverse weather such as rain and fog, similar target appearance, and drastic scale variation of targets. Conventional frame-based cameras capture visual scenes at a fixed frame rate and rely primarily on texture, color, and illumination cues. As a result, frame-based trackers often suffer from significant performance degradation under adverse conditions, such as low illumination, overexposure, and motion blur. These real-world challenges are particularly detrimental to frame-only MOT systems, as unreliable object perception can lead to fragmented trajectories and accumulated association errors over time. Therefore, relying solely on frame-based visual signals inherently limits the robustness of MOT systems in practical applications.

\begin{figure*}
    \centering
    \includegraphics[width=1\linewidth]{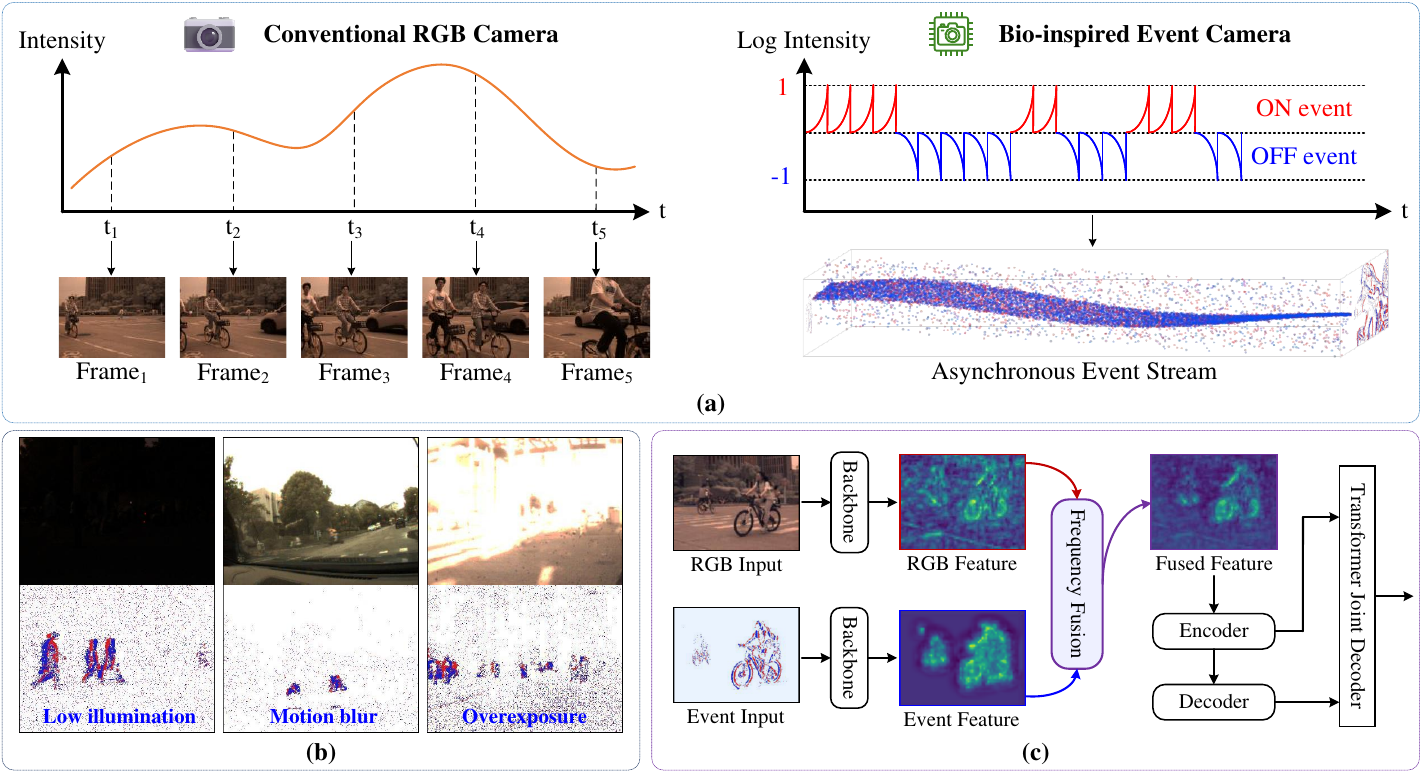}
    \caption{(a) Comparison of imaging mechanisms between conventional RGB and bio-inspired event cameras; (b) Advantages of event cameras under extreme conditions; (c) Schematic illustration of the proposed framework. }
    \label{fig:First_Image}
\end{figure*}

Bio-inspired event cameras offer a promising complementary sensing modality for addressing these limitations. Unlike conventional cameras, which capture intensity images at fixed intervals, event cameras asynchronously record pixel-level brightness changes with microsecond-level temporal resolution, as shown in Fig.~\ref{fig:First_Image} (a). This unique sensing mechanism provides several attractive properties, such as high dynamic range, low latency, and reduced motion blur, enabling robust perception of fast motion and challenging illumination conditions~\cite{gallego2020event}, as shown in Fig.~\ref{fig:First_Image} (b). When integrated with conventional frame cameras, event data can provide rich motion-sensitive and high-frequency cues, whereas frame images retain dense appearance and semantic information. This inherent complementarity makes frame-event perception a natural and promising paradigm for robust MOT.

Despite recent efforts, such as DSEC-MOT~\cite{wang2024spikemot}, which represents important progress in event-based MOT benchmarking, RGB-event MOT has not yet been systematically investigated. Existing studies have demonstrated the potential of RGB-event perception in various tasks, including object detection~\cite{cui2026peod}, person re-identification~\cite{wang2026evreid}, and single-object tracking~\cite{wang2024visevent}. However, MOT is inherently more challenging than these tasks, since it requires not only accurate object localization in individual frames but also temporally consistent identity association across frames. This requirement places higher demands on annotation granularity, especially identity-level trajectory annotations. As a result, the field still lacks a comprehensive, diverse, and well-annotated RGB-event MOT benchmark dataset.

\begin{table*}[!t]
\centering
\caption{Comparison between existing MOT datasets, RGB-event SOT datasets, and our FEMOT dataset.}
\label{tab:dataset_comparison}
\resizebox{\textwidth}{!}{
\begin{tabular}{c|l|c|c|c|c|c|c|c|c}
\hline
\textbf{No.} & \textbf{Dataset} & \textbf{Year} & \textbf{Venue} & \textbf{Modality} & \textbf{Task} 
& \textbf{\# Videos} & \textbf{\# Frames} & \textbf{\# Tracks} & \textbf{\# Ann. Boxes} \\
\hline
01 & MOT16~\cite{milan2016mot16}        & 2016 & MOTChallenge & RGB          & MOT & 14    & 11.24K & 1.34K   & 0.30M \\
02 & MOT17~\cite{milan2016mot16}        & 2016 & MOTChallenge & RGB          & MOT & 42    & 33.71K & 3.99K   & 0.90M \\
03 & MOT20~\cite{dendorfer2020mot20}    & 2020 & CVPR         & RGB          & MOT & 8     & 13.40K & 3.83K   & 2.10M \\
04 & DanceTrack~\cite{sun2022dancetrack}& 2022 & CVPR         & RGB          & MOT & 100   & 105.86K & 0.99K  & 0.57M \\
05 & SportsMOT~\cite{cui2023sportsmot}  & 2023 & ICCV         & RGB          & MOT & 240   & 150.38K & 3.40K  & 1.63M \\
06 & TAO~\cite{dave2020tao}             & 2020 & ECCV         & RGB          & MOT & 2,907 & 4.45M  & 16.10K  & 0.33M \\
07 & BDD100K~\cite{yu2020bdd100k}       & 2018 & CVPR         & RGB          & MOT & 2,000    & 318K   & 130.60K & 3.30M \\
\hline
08 & VTMOT~\cite{zhu2025visible}        & 2025 & PR           & RGB-Infrared & MOT & 582   & 401K   & 13K     & 4.00M \\
\hline
09 & FE108~\cite{Zhang_2021_ICCV}       & 2021 & ICCV         & RGB-Event    & SOT & 108   & 209K   & -      & 0.21M \\
10 & VisEvent~\cite{wang2024visevent}   & 2024 & TCYB         & RGB-Event    & SOT & 820   & 371K   & -      & 0.37M \\
11 & COESOT~\cite{tang2025revisiting}   & 2025 & PR           & RGB-Event    & SOT & 1,354 & 479K   & -      & 0.48M \\
\hline
12 & DSEC-MOT~\cite{wang2024spikemot}   & 2025 & IEEE Access  & RGB-Event    & MOT & 12    & 23.08K & 0.50K   & 0.037M \\
13 & FEMOT (Ours)              & 2026 & -           & RGB-Event    & MOT & 100   & 200K   & 14.58K  & 0.42M \\
\hline
\end{tabular}
}
\end{table*}

To bridge this dataset gap, we propose FEMOT, a new large-scale RGB-event multimodal dataset for multi-object tracking. Unlike existing MOT datasets reported in Table~\ref{tab:dataset_comparison}, FEMOT is carefully designed to cover diverse real-world challenges, including 14 attributes such as low illumination, fast motion, and frequent target entry and exit, thereby enabling comprehensive evaluation of tracker robustness in complex environments. The sequences in FEMOT are captured using a DVS346 camera, which provides spatially well-aligned frame and event streams. FEMOT mainly focuses on two common object categories in real-world scenarios, namely people and vehicles, and contains 100 video sequences, 200K video frames, 14.58K trajectories, and 0.42M bounding box annotations. To establish a comprehensive benchmark platform, we further report comparative results of multiple state-of-the-art trackers, providing a solid benchmark for future research. We hope our newly proposed FEMOT dataset can open up new possibilities for RGB-event-based MOT research.

Based on the newly proposed FEMOT dataset, we further propose a new \textbf{F}requency-aware RGB Frame-\textbf{E}vent \textbf{M}ulti-\textbf{O}bject Tracking method with \textbf{TR}ansformer, termed \textbf{FEMOTR}. Different from conventional spatial-domain fusion methods, FEMOTR exploits the complementary frequency characteristics of RGB frames and event streams, where RGB frames mainly provide low-frequency appearance and semantic cues, while event streams capture high-frequency motion and contour information. Specifically, RGB, event, and concatenated fusion features are individually decomposed into amplitude and phase components in the frequency domain, where the fused components modulate the corresponding RGB and event branches through an amplitude/phase attention module, followed by the integration of the three spatial-domain branches to obtain the enhanced multimodal representation. The enhanced features are then fed into an encoder-decoder architecture, where a dynamic temporal interaction module performs memory-based association to predict the locations and identities of multiple targets in the current scene. A schematic illustration of the proposed FEMOTR can be found in Fig.~\ref{fig:First_Image} (c).

To sum up, the main contributions of this work can be summarized as follows:

$\bullet$ We introduce FEMOT, a new large-scale RGB-event multimodal dataset for multi-object tracking. We also provide experimental evaluations of recent
strong trackers to build a comprehensive RGB-event-based tracking benchmark.

$\bullet$ We propose FEMOTR, a new RGB-event MOT framework that exploits the complementary frequency characteristics of RGB frames and event streams.

$\bullet$ Extensive experiments on two benchmark datasets, i.e., DSEC-MOT and FEMOT, fully validate the effectiveness of the proposed tracker.


\section{Related Work}

\subsection{Multi-Object Tracking} 

Existing multiple object tracking (MOT) methods can be broadly categorized into three representative paradigms: tracking-by-detection, joint detection-and-tracking, and query-based Transformer tracking~\cite{chen2024delving}. 
Among them, tracking-by-detection has long been the dominant framework due to its simplicity, flexibility, and strong empirical performance.

Tracking-by-detection methods typically decompose MOT into two stages: object detection and data association. 
Given detections in the current frame and tracklets maintained from previous frames, the key problem is to construct an affinity matrix and solve the resulting bipartite matching problem, commonly using the Hungarian algorithm~\cite{kuhn1955hungarian}. 
The affinity matrix is usually built upon motion cues~\cite{qin2023motiontrack, han2022mat}, appearance cues~\cite{pang2021quasi, wang2021multiple, wojke2017simple}, or their combination. 
To exploit the complementary strengths of motion and appearance information, many methods~\cite{aharon2022bot, du2023strongsort, zhang2022bytetrack} design effective association strategies to improve tracking accuracy while maintaining a favorable balance between robustness and computational complexity~\cite{qin2024towards}.

Joint detection-and-tracking methods~\cite{bergmann2019tracking, liang2022rethinking, wang2020towards} aim to reduce the separation between detection and association by learning both tasks within a unified framework. 
Instead of relying on independently trained detectors and separate ReID models, these methods jointly predict object detections and association-related cues, such as appearance embeddings, motion offsets, or object center displacements. 
The predicted association cues are then used to link objects across consecutive frames, resulting in a more compact and efficient tracking pipeline.

More recently, query-based Transformer tracking methods~\cite{zhou2022global,sun2020transtrack,zeng2022motr,meinhardt2022trackformer,gao2023memotr} have formulated tracking as a query propagation problem. 
By introducing detection queries or track queries, these methods maintain object identities through temporal query updates and attention-based feature interaction. 
Compared with conventional tracking-by-detection pipelines, query-based Transformer trackers reduce the reliance on hand-crafted association rules and external ReID modules, providing a more end-to-end solution for MOT.
Following this paradigm, our work builds an RGB-event MOT framework that maintains object identities through query propagation while exploiting complementary cues from RGB frames and event streams.


\subsection{Benchmark Datasets for MOT}

Benchmark datasets play a crucial role in driving the development of the MOT task. The MOT Challenge benchmarks, including MOT15~\cite{leal2015motchallenge}, MOT17~\cite{milan2016mot16}, and MOT20~\cite{dendorfer2020mot20}, have been widely used to evaluate pedestrian tracking methods under increasingly challenging scenarios. By providing standardized annotations, evaluation protocols, and diverse video sequences, these benchmarks have enabled fair comparison among MOT methods. 
BDD100K~\cite{yu2020bdd100k} is a large-scale driving video dataset designed for autonomous driving perception. It contains diverse scenarios across different locations, weather conditions, and times of day, and provides annotations for multiple perception tasks, including object detection, tracking, segmentation, and lane detection. For MOT, BDD100K offers a challenging benchmark in real-world driving environments, where trackers must handle scale variation, camera motion, occlusion, and complex traffic participants.

Beyond general pedestrian and driving-scene benchmarks, several datasets have been proposed to investigate more challenging motion patterns. DanceTrack~\cite{sun2022dancetrack} is designed to address the difficulties caused by similar appearances, diverse poses, and highly non-linear motion. Consisting mainly of group dance videos, it encourages the development of MOT algorithms that rely more on motion modeling rather than appearance discrimination. Similarly, SportsMOT~\cite{cui2023sportsmot} focuses on large-scale sports scenarios, which are characterized by fast motion, frequent interactions, and visually similar targets. It aims to facilitate the development of trackers that effectively exploit both motion and appearance cues.

Recently, multi-modal MOT datasets have attracted increasing attention. 
VT-MOT~\cite{zhu2025visible} provides a large-scale visible-thermal MOT benchmark, encouraging the exploration of complementary cues from RGB and thermal infrared modalities for robust tracking. 
Built upon DSEC~\cite{Gehrig21ral}, DSEC-MOT~\cite{wang2024spikemot} provides tracking annotations for challenging driving scenarios with RGB frames and event streams. However, its original benchmark setting mainly focuses on event-based MOT, while the systematic evaluation of cross-modal complementarity between RGB frames and event streams remains insufficiently explored. 
Therefore, RGB-event MOT still lacks a widely adopted benchmark across diverse scenarios. 
To bridge this gap, we propose a novel multi-scenario RGB-event MOT dataset to establish a comprehensive benchmark for MOT. This dataset facilitates the development of robust trackers by exploiting event data as a complement to RGB cameras in challenging scenarios.

\subsection{RGB-Event based Fusion}

RGB-event fusion has emerged as a promising direction for robust visual perception by exploiting the complementary characteristics of frame-based and event-based sensors. 
Recent studies have explored RGB-event fusion across various visual tasks, including object detection, semantic segmentation, and single object tracking (SOT). 
For robust object detection, Tomy et al.~\cite{tomy2022fusing} convert asynchronous event streams into frame-like representations and fuse them with RGB features at multiple network stages, demonstrating improved robustness under degraded illumination and motion blur. 
In autonomous driving scenarios, RGB-event fusion has also been investigated for moving object detection, where event streams provide complementary motion cues to RGB frames for perceiving dynamic objects~\cite{zhou2023rgb}.
For the SOT task, Tang et al.~\cite{tang2025revisiting} introduce a unified Transformer-based framework that jointly performs feature extraction, cross-modal fusion, and matching.
More recent methods~\cite{zhou2024bring,gehrig2024low,liu2025beyond} further enhance multimodal interaction through adaptive fusion modules, hierarchical visual-motion fusion, or spatiotemporal alignment mechanisms, highlighting the effectiveness of the event modality in improving RGB-based perception under challenging environments. 
These studies suggest that event cameras can serve as a strong complementary sensor for RGB cameras, particularly when conventional frame-based perception becomes unreliable.

Despite recent progress, RGB-event fusion remains underexplored in the MOT task, where the fused representation needs to support not only object localization but also reliable identity association under occlusion and inter-object interactions. 
In this work, we propose a frequency-aware feature fusion mechanism that explicitly models modality-specific frequency characteristics and enhances complementary RGB-event frequency features for MOT.

\section{Methodology}

\subsection{Overview}  

\begin{figure*}
    \centering
    \includegraphics[width=1\linewidth]{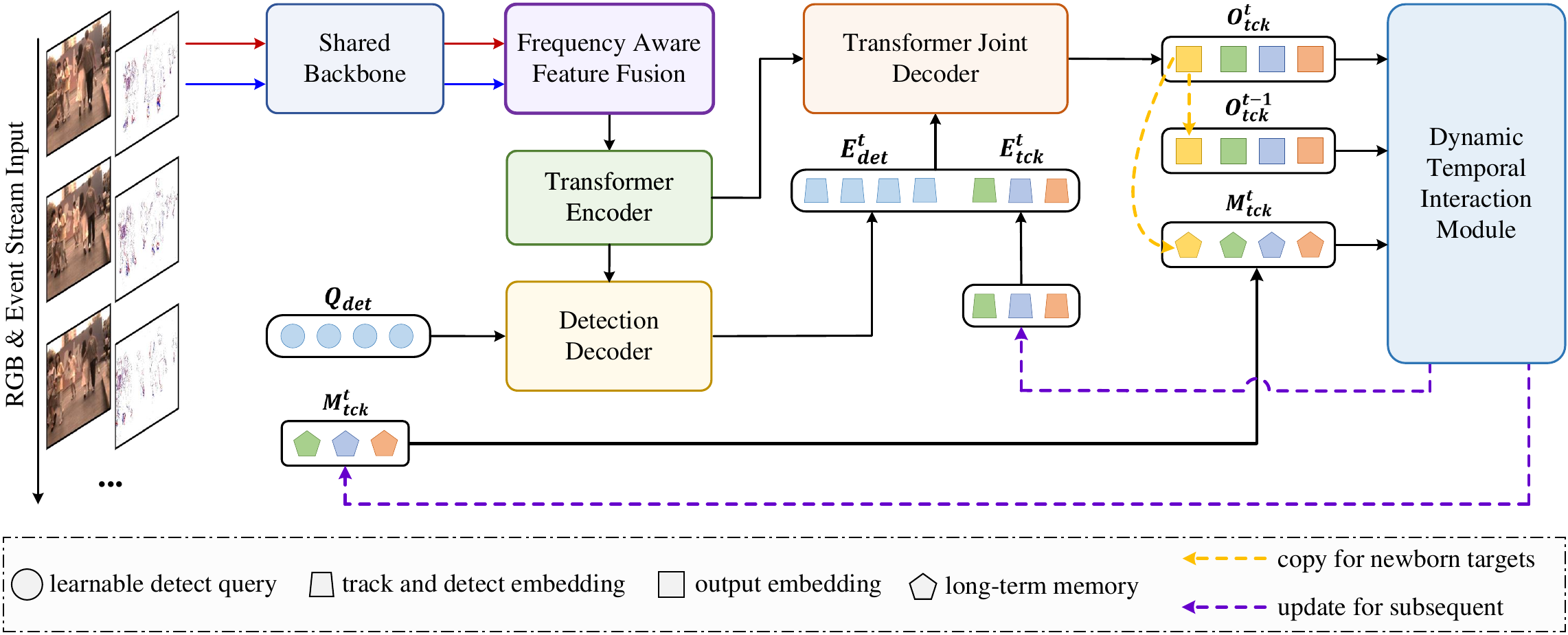}
    \caption{A multi-object tracking framework based on frequency-aware feature fusion with RGB images and event streams. Through the shared backbone network, the RGB and event features are fed into the frequency-aware feature fusion module, where the low-frequency semantic information from the RGB modality and the high-frequency contour information from the event modality are fully exploited and integrated. Finally, the fused features are input into the encoder-decoder architecture for object localization and identity association, while the dynamic temporal interaction module is employed to maintain long-term memory.
}
    \label{fig:framework}
\end{figure*}

As illustrated in Fig.~\ref{fig:framework}, the RGB and event streams are first fed into a shared backbone to extract multi-scale features, which are then enhanced by the Frequency Aware Feature Fusion module. The fused features are further processed by a Transformer Encoder and serve as the common feature memory for subsequent decoding. In the detection branch, a set of learnable detection queries (300 by default) is fed into the Detection Decoder and interacts with the encoded fused features to generate the detection embeddings. Meanwhile, the track embeddings, propagated from the track memory of previous frames, are delivered to the Transformer Joint Decoder together with the detection embeddings. The Joint Decoder jointly models detection and tracking embeddings to associate newly detected targets with existing trajectories and to produce updated tracking outputs. Specifically, its outputs are used to generate current-frame track predictions, copy representations for newly born targets, and update the track memory. The updated memory is then fed into the Dynamic Temporal Interaction Module (DTIM), which maintains current and historical track queries as well as long-term temporal memory for subsequent frames. During training, track queries are assigned to ground-truth objects through Hungarian-algorithm-based bipartite matching for loss computation. During inference, the updated track queries are directly used for target prediction.

\subsection{Input Representation}  
Given an RGB video sequence with $N$ frames, denoted as 
$\mathcal{I} = \{\mathcal{I}^k\}_{k=1}^{N}$, where 
$\mathcal{I}^k \in \mathbb{R}^{3 \times H \times W}$ and $H$ and $W$ denote the spatial resolution, the corresponding asynchronous event stream is represented as 
$\mathcal{E}_p = \{e_i\}_{i=1}^{M}$. Each event $e_i$ is defined as a quadruple 
$(x_i, y_i, t_i, p_i)$, where $(x_i, y_i)$ denotes the spatial coordinates, 
$t_i \in [0, T]$ denotes the timestamp, and $p_i$ denotes the polarity. Here, $M$ is the total number of events in the event stream. 

To make event data compatible with prevailing deep learning models and temporally aligned with RGB frames, we convert the asynchronous event stream into frame-like representations according to the temporal interval of each RGB frame. Specifically, for the $t$-th RGB frame $\mathcal{I}^t \in \mathbb{R}^{3 \times H \times W}$, we use its exposure start and end timestamps, denoted as $t_s$ and $t_e$, to define the event accumulation window. The corresponding event segment is given by
\begin{equation}
\mathcal{E}_p^t =
\left\{
e_i = (x_i, y_i, t_i, p_i)
\mid
t_i \in [t_s, t_e]
\right\}.
\label{eq:event_segment}
\end{equation}
The selected events are then projected onto the image plane according to their spatial coordinates. ON and OFF events are encoded with different colors on a three-channel canvas, yielding the event frame 
$\mathcal{E}^t \in \mathbb{R}^{3 \times H \times W}$.

\subsection{Network Architecture} 

As shown in Fig.~\ref{fig:framework}, our framework takes the RGB frame $\mathcal{I}^t$ and the event image $\mathcal{E}^t$ at the current timestep as inputs. We first employ a shared ResNet-50~\cite{he2016deep} backbone to extract multi-scale hierarchical features from the two modalities. The RGB features are denoted as $X^t=\{X_l^t\}_{l=1}^{L}$, where $X_l^t$ represents the feature map from the $l$-th backbone stage. Similarly, the corresponding event features are denoted as $Y^t=\{Y_l^t\}_{l=1}^{L}$.
Subsequently, the multi-scale RGB and event features are fed into the frequency-aware feature fusion module, which produces enhanced fused features $F^t=\{F_l^t\}_{l=1}^{L}$. The fused features are then flattened, augmented with positional and level embeddings, and processed by a multi-scale deformable Transformer encoder to obtain the encoded representations $F_{\mathrm{en}}^t$.

The decoding process is conceptually divided into two stages. In the first stage, a set of learnable detection queries $Q_{\mathrm{det}}$ interacts with the encoded features $F_{\mathrm{en}}^t$ through the detection decoding stage, yielding dynamic detection embeddings $E_{\mathrm{det}}^t$. In the second stage, the dynamic detection embeddings are concatenated with the propagated track embeddings $E_{\mathrm{tck}}^t$ from previous frames and jointly decoded by the Transformer joint decoder. This joint decoding stage enables interaction between newly detected candidates and existing tracks, producing the output embeddings $[O_{\mathrm{det}}^t, O_{\mathrm{tck}}^t]$.
The output embeddings are further fed into the prediction heads to estimate the classification confidence $c_i^t$ and bounding box $b_i^t$ for each target. Specifically, $O_{\mathrm{det}}^t$ is mainly responsible for discovering newly appearing objects in the current frame, while $O_{\mathrm{tck}}^t$ corresponds to targets propagated from previous timesteps.

To maintain temporal consistency and exploit historical information, the output track embeddings $O_{\mathrm{tck}}^t$, together with the previous output embeddings $O_{\mathrm{tck}}^{t-1}$ and the long-term memory embeddings $M_{\mathrm{tck}}^t$, are passed into the Dynamic Temporal Interaction Module. This module updates the track embeddings and long-term memory, producing $E_{\mathrm{tck}}^{t+1}$ and $M_{\mathrm{tck}}^{t+1}$ for the next frame. In this way, the framework propagates reliable target representations over time and achieves robust multi-object tracking in long video sequences.

\subsection{Frequency Aware Feature Fusion Module}

\begin{figure}
    \centering
    \includegraphics[width=0.98\linewidth]{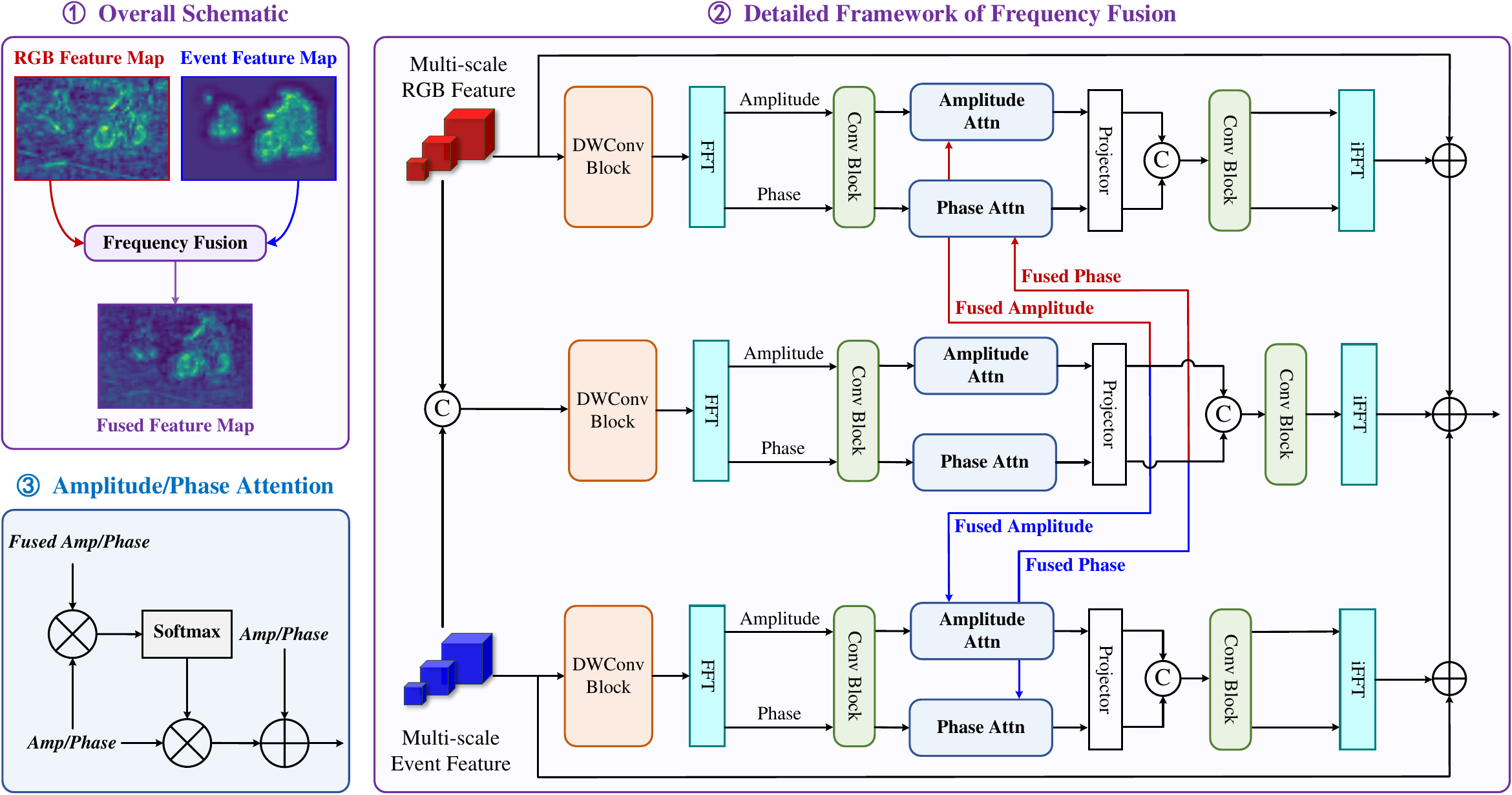}
    \caption{Illustration of Frequency Aware Feature Fusion Module.}
    \label{fig:FAF}
\end{figure}

RGB features usually provide rich appearance and semantic information, while event features are more sensitive to motion boundaries and rapid intensity changes. Direct spatial-domain fusion may insufficiently exploit such complementary frequency patterns and may be vulnerable to modality-specific noise. Therefore, we introduce a Frequency Aware Feature Fusion (FAF) module to enhance cross-modal representations.


As illustrated in Fig.~\ref{fig:FAF}, given the multi-scale RGB features 
$X^t=\{X_l^t\}_{l=1}^{L}$ and event features 
$Y^t=\{Y_l^t\}_{l=1}^{L}$, the proposed FAF module performs cross-modal interaction in the frequency domain. Since the FAF module is independently applied to each feature scale, the scale index $l$ is omitted in the following description for notational simplicity.

The FAF module consists of three branches: an RGB branch, an event branch, and a cross-modal fusion branch. The inputs to these branches are defined as
\begin{equation}
    B_X^t = X^t, \qquad
    B_Y^t = Y^t, \qquad
    B_Z^t = Z^t = \operatorname{Concat}(X^t,Y^t).
\end{equation}
For each branch $b \in \{X, Y, Z\}$, a depthwise separable convolution block is first employed to capture local spatial patterns:
\begin{equation}
    U_b^t = \operatorname{DWConv}_b(B_b^t).
\end{equation}
The resulting features are then transformed into the frequency domain by a two-dimensional fast Fourier transform (FFT):
\begin{equation}
    S_b = \mathcal{F}(U_b^t),
\end{equation}
where $\mathcal{F}(\cdot)$ denotes the 2D FFT. The amplitude and phase components are further decoupled and processed by channel convolution layers:
\begin{equation}
    A_b = \operatorname{Conv}_b\bigl(\mathcal{A}(S_b)\bigr), \qquad
    P_b = \operatorname{Conv}_b\bigl(\mathcal{P}(S_b)\bigr),
\end{equation}
where $\mathcal{A}(\cdot)$ and $\mathcal{P}(\cdot)$ denote amplitude and phase extraction, respectively.

The cross-modal fusion branch is designed to generate adaptive frequency guidance for the two modality-specific branches. Given a frequency component $V$ and its corresponding guidance component $W$, the modulation operation is formulated as
\begin{equation}
    \operatorname{Attn}(V,W)
    =
    V + V \odot \operatorname{Softmax}(V \odot W),
\end{equation}
where $\odot$ denotes element-wise multiplication.
We first enhance the frequency representation of the fusion branch through self-modulation:
\begin{equation}
    \tilde{A}_Z
    =
    \operatorname{Proj}_Z
    \bigl(
    \operatorname{Attn}(A_Z,A_Z)
    \bigr),
    \qquad
    \tilde{P}_Z
    =
    \operatorname{Proj}_Z
    \bigl(
    \operatorname{Attn}(P_Z,P_Z)
    \bigr).
\end{equation}
Here, $\tilde{A}_Z$ and $\tilde{P}_Z$ serve as cross-modal guidance for amplitude and phase modulation, respectively. Guided by these representations, the RGB and event branches are enhanced as
\begin{equation}
    \tilde{A}_b
    =
    \operatorname{Proj}_{b}
    \bigl(
    \operatorname{Attn}(A_b,\tilde{A}_Z)
    \bigr),
    \qquad
    \tilde{P}_b
    =
    \operatorname{Proj}_{b}
    \bigl(
    \operatorname{Attn}(P_b,\tilde{P}_Z)
    \bigr),
    \quad b \in \{X,Y\}.
\end{equation}
In this manner, the fusion branch emphasizes frequency responses shared across modalities, while the RGB and event branches preserve their modality-specific characteristics.

After modulation, the enhanced amplitude and phase components are recombined to reconstruct the complex-valued frequency representation of each branch:
\begin{equation}
    \tilde{S}_b
    =
    \tilde{A}_b \odot e^{j\tilde{P}_b},
    \qquad b \in \{X,Y,Z\}.
\end{equation}
A convolution block is then applied to integrate the enhanced frequency-domain information. The resulting representation is transformed back to the spatial domain through inverse FFT, with a residual connection from the corresponding spatial feature:
\begin{equation}
    \hat{B}_b^t
    =
    \mathcal{F}^{-1}
    \bigl(
    \operatorname{Conv}_b(\tilde{S}_b)
    \bigr)
    + B_b^t,
    \qquad b \in \{X,Y,Z\},
\end{equation}
where $\mathcal{F}^{-1}(\cdot)$ denotes the inverse FFT. For clarity, the reconstructed outputs of the three branches are denoted as
\begin{equation}
    \hat{X}^t = \hat{B}_X^t, \qquad
    \hat{Y}^t = \hat{B}_Y^t, \qquad
    \hat{Z}^t = \hat{B}_Z^t.
\end{equation}

Finally, the enhanced RGB, event, and cross-modal fusion features are aggregated by element-wise addition to obtain the fused representation:
\begin{equation}
    F^t = \hat{X}^t + \hat{Y}^t + \hat{Z}^t.
\end{equation}
The resulting feature $F^t$ integrates modality-specific information and complementary cross-modal frequency cues, providing a more discriminative representation for subsequent detection and tracking.

\begin{figure}
    \centering
    \includegraphics[width=0.95\linewidth]{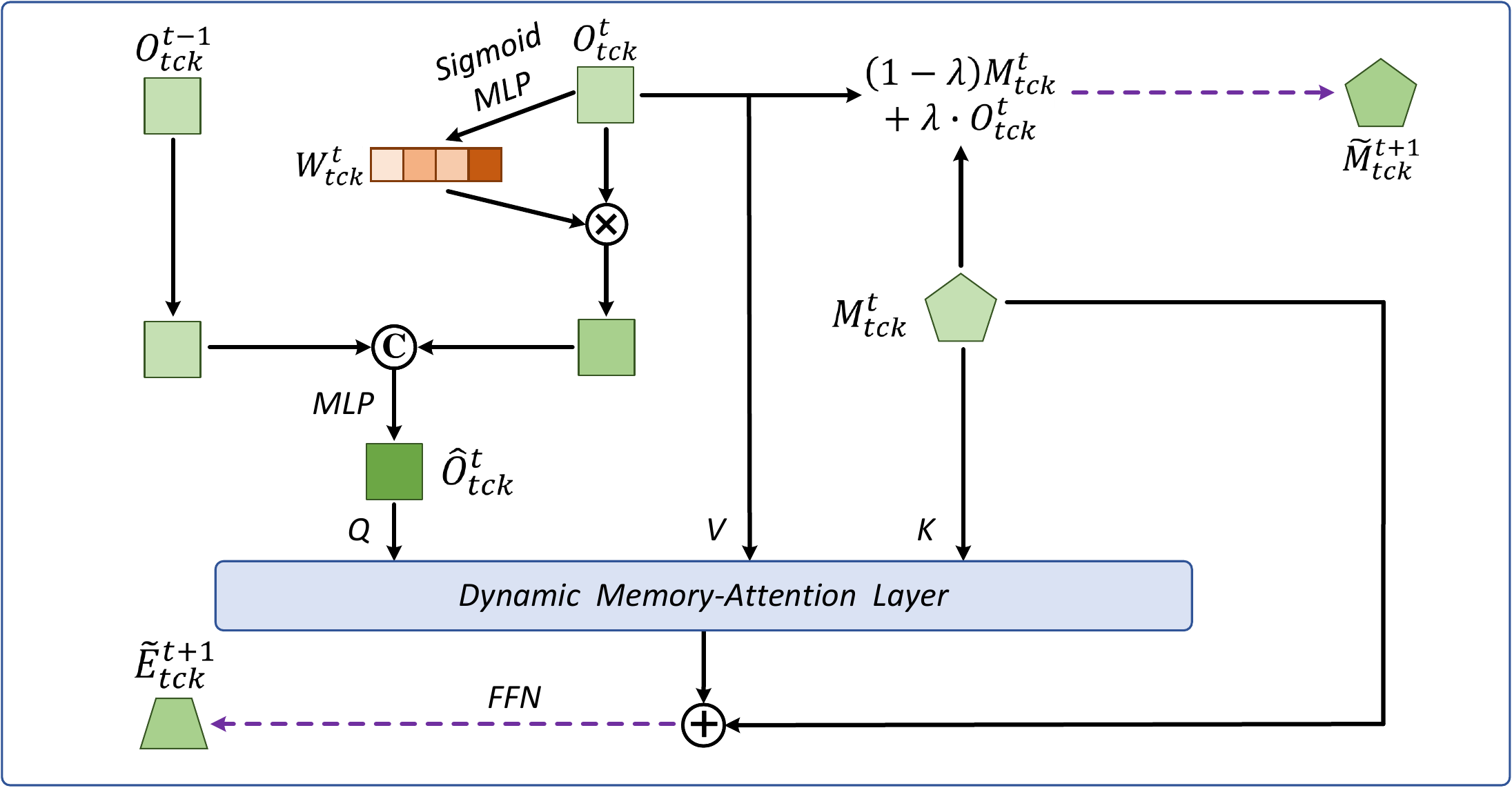}
    \caption{Illustration of Dynamic Temporal Interaction Module.}
    \label{fig:DTIM}
\end{figure}

\subsection{Dynamic Temporal Interaction Module}

To achieve robust long-term tracking, we follow MeMOTR~\cite{gao2023memotr} and incorporate a dynamic temporal interaction module. As shown in Fig.~\ref{fig:DTIM}, this module takes the output embeddings $O_{\mathrm{tck}}^{t-1}$ and $O_{\mathrm{tck}}^{t}$ from the previous and current frames, respectively, together with the long-term memory $M_{\mathrm{tck}}^{t}$ as inputs. It produces the updated long-term memory $\widetilde{M}_{\mathrm{tck}}^{t+1}$, as well as the tracking and detection embedding $\widetilde{E}_{\mathrm{tck}}^{t+1}$ for the next frame.

The long-term memory is updated using the current output embedding. Specifically, we employ an exponential moving average strategy, which enables the long-term memory to mainly preserve historical information while gradually absorbing current-frame information. In this way, a more stable tracking representation can be obtained. The update process is formulated as follows:
\begin{equation}
\widetilde{M}_{\mathrm{tck}}^{t+1} =
(1-\lambda)M_{\mathrm{tck}}^{t} + \lambda \cdot O_{\mathrm{tck}}^{t},
\end{equation}
where $\widetilde{M}_{\mathrm{tck}}^{t+1}$ denotes the updated long-term memory for the next frame. We empirically set the memory update rate $\lambda$ to $0.01$ to ensure smooth and consistent memory evolution across consecutive frames. In particular, when a newly detected object appears, its long-term memory is initialized with the corresponding current output embedding.

To obtain the track embeddings $\widetilde{E}_{\mathrm{tck}}^{t+1}$ for the next frame, we first generate channel-wise weights $W_{\mathrm{tck}}^{t}$ using an MLP followed by a Sigmoid activation function. These weights adaptively emphasize reliable channels and suppress unreliable ones in the current-frame output embedding $O_{\mathrm{tck}}^{t}$. We then multiply $W_{\mathrm{tck}}^{t}$ with $O_{\mathrm{tck}}^{t}$ and concatenate the result with the output embedding from the previous frame $O_{\mathrm{tck}}^{t-1}$. Finally, a two-layer MLP is employed to produce the fused output embedding $\hat{O}_{\mathrm{tck}}^{t}$. This process can be formulated as follows:
\begin{equation}
W_{\mathrm{tck}}^{t} =
\operatorname{Sigmoid}
\left(
\operatorname{MLP}_{1}
\left(
O_{\mathrm{tck}}^{t}
\right)
\right),
\end{equation}
\begin{equation}
\hat{O}_{\mathrm{tck}}^{t} =
\operatorname{MLP}_{2}
\left(
\left[
W_{\mathrm{tck}}^{t} \odot O_{\mathrm{tck}}^{t},
O_{\mathrm{tck}}^{t-1}
\right]
\right),
\end{equation}
where $\odot$ denotes channel-wise multiplication, $[\cdot,\cdot]$ denotes concatenation, and $\hat{O}_{\mathrm{tck}}^{t}$ is the fused output embedding.

Subsequently, we employ a dynamic memory-attention layer based on multi-head attention to enable effective interactions among different trajectories. Specifically, the fused output embedding $\hat{O}_{\mathrm{tck}}^{t}$, the current long-term memory $M_{\mathrm{tck}}^{t}$, and the current output embedding $O_{\mathrm{tck}}^{t}$ are used as the query, key, and value, respectively. The output of the dynamic memory-attention layer is then added to the long-term memory $M_{\mathrm{tck}}^{t}$, and further passed through an FFN to predict the track embedding $\widetilde{E}_{\mathrm{tck}}^{t+1}$ for the next frame.

\subsection{Loss Function} 
Our framework is optimized with a multi-task objective consisting of a classification loss and two box localization losses. Specifically, we adopt the sigmoid Focal Loss to alleviate class imbalance by reducing the contribution of easy samples and emphasizing hard examples. For bounding box regression, the L1 loss is used to measure coordinate-wise localization errors, while the GIoU loss imposes an overlap- and geometry-aware constraint by considering both the IoU between the predicted and ground-truth boxes and their smallest enclosing box. The total loss is formulated as:
\begin{equation}
\mathcal{L}_{\mathrm{total}} =
\lambda_1\mathcal{L}_{\mathrm{Focal}} +
\lambda_2 \mathcal{L}_{\mathrm{L1}} +
\lambda_3 \mathcal{L}_{\mathrm{GIoU}},
\end{equation}
where the loss weights are set to
$\lambda_1 = 2$,
$\lambda_2 = 5$, and
$\lambda_3 = 2$, respectively.

During training, Hungarian matching is adopted to solve a minimum-cost bipartite assignment problem between detection predictions and the corresponding ground-truth targets. The matching cost is defined as a weighted sum of three terms:
\begin{equation}
\mathcal{C}_{ij} =
\lambda_4 \mathcal{C}_{\mathrm{cls}}(i,j) +
\lambda_5 \mathcal{C}_{\mathrm{L1}}(i,j) +
\lambda_6 \mathcal{C}_{\mathrm{giou}}(i,j),
\end{equation}
where $\mathcal{C}_{\mathrm{cls}}$, $\mathcal{C}_{\mathrm{L1}}$, and $\mathcal{C}_{\mathrm{giou}}$ denote the classification, L1 bounding box, and GIoU matching costs, respectively. The corresponding weights are set to $\lambda_4=2$, $\lambda_5=5$, and $\lambda_6=2$, respectively. In the tracking setting, detection queries are matched to ground-truth objects that are not already associated with existing track queries, while propagated track queries are supervised according to their associated identities. The matched prediction-target pairs are then used to compute the classification and localization losses.

\section{FEMOT Benchmark Dataset} 
In this paper, we introduce a comprehensive, large-scale dataset for RGB-event-based multiple object tracking across diverse scenes, called FEMOT. In the following subsections, we describe the data collection and annotation protocols, provide statistical analysis, and present the benchmarked trackers. Representative samples are shown in Fig.~\ref{fig:dataset_examples} (b).

\subsection{Protocols} 
To build a comprehensive and challenging benchmark, we carefully considered several key factors during data collection, aiming to ensure that the resulting dataset reflects both the distinctive characteristics of RGB-event sensing and the practical challenges encountered in real-world tracking scenarios.
These factors are summarized as follows.
\textit{1) Diversity of data collection environments.}
The dataset contains video sequences captured under a wide range of environmental conditions, including both daytime and nighttime scenes. The recording locations were carefully selected to cover diverse real-world scenarios, such as school entrances, canteen interiors and exteriors, indoor and outdoor teaching buildings, shopping mall surroundings, urban streets, and overpass bridges. Such environmental diversity improves the applicability of the dataset to various real-world tracking scenarios.
\textit{2) Diversity of camera motion states.}
During data acquisition, the camera exhibits complex motion patterns beyond the static settings commonly seen in conventional tracking datasets. Specifically, the dataset includes irregular rotations, translational movements, and abrupt camera shaking. In addition, it covers diverse viewpoints, including both eye-level and bird's-eye perspectives. To further enhance the generalization ability and practical relevance of the dataset, we also include vehicle-mounted camera scenarios. These dynamic camera motion patterns significantly increase the difficulty of multi-object tracking, especially in maintaining identity consistency and localization accuracy under motion-induced disturbances.
\textit{3) Scenarios tailored to event-camera characteristics.}
The dataset is specifically designed to reflect the unique sensing properties of event cameras. It includes scenarios with different motion speeds, such as high-speed motion, slow motion, and momentary stillness, as well as diverse illumination conditions. These settings enable a comprehensive evaluation of event-based tracking algorithms under realistic and challenging conditions.
\textit{4) High-quality data annotation.}
All annotations were produced using the open-source tool X-AnyLabeling~\cite{X-AnyLabeling}. To ensure annotation reliability, we conducted multiple rounds of verification and iterative refinement. In addition, each sequence was adaptively processed to retain approximately 1,000 frames, taking into account variations in scene content and illumination conditions.
\textit{5) Dataset scale and robustness.}
The dataset is constructed with a sufficient scale to support the training and evaluation of robust RGB-event-based trackers. A detailed comparison between the proposed dataset and existing tracking datasets is provided in Table~\ref{tab:dataset_comparison}, highlighting the distinctive characteristics, challenges, and advantages of our dataset.

\subsection{Annotations}

To ensure high-quality annotations for the MOT dataset, we adopted a rigorous annotation protocol. The annotation process was carried out by a team of trained volunteers under a strict quality control mechanism. After the initial annotation stage, multiple rounds of careful verification and iterative refinement were conducted to identify potential errors and improve annotation accuracy. The annotation process was facilitated by the open-source software X-AnyLabeling~\cite{X-AnyLabeling}, which provides an efficient and user-friendly annotation platform. The entire annotation process lasted several months, during which repeated quality checks were performed to ensure the reliability and precision of the final annotations.

\begin{figure*}
    \centering
    \includegraphics[width=\linewidth]{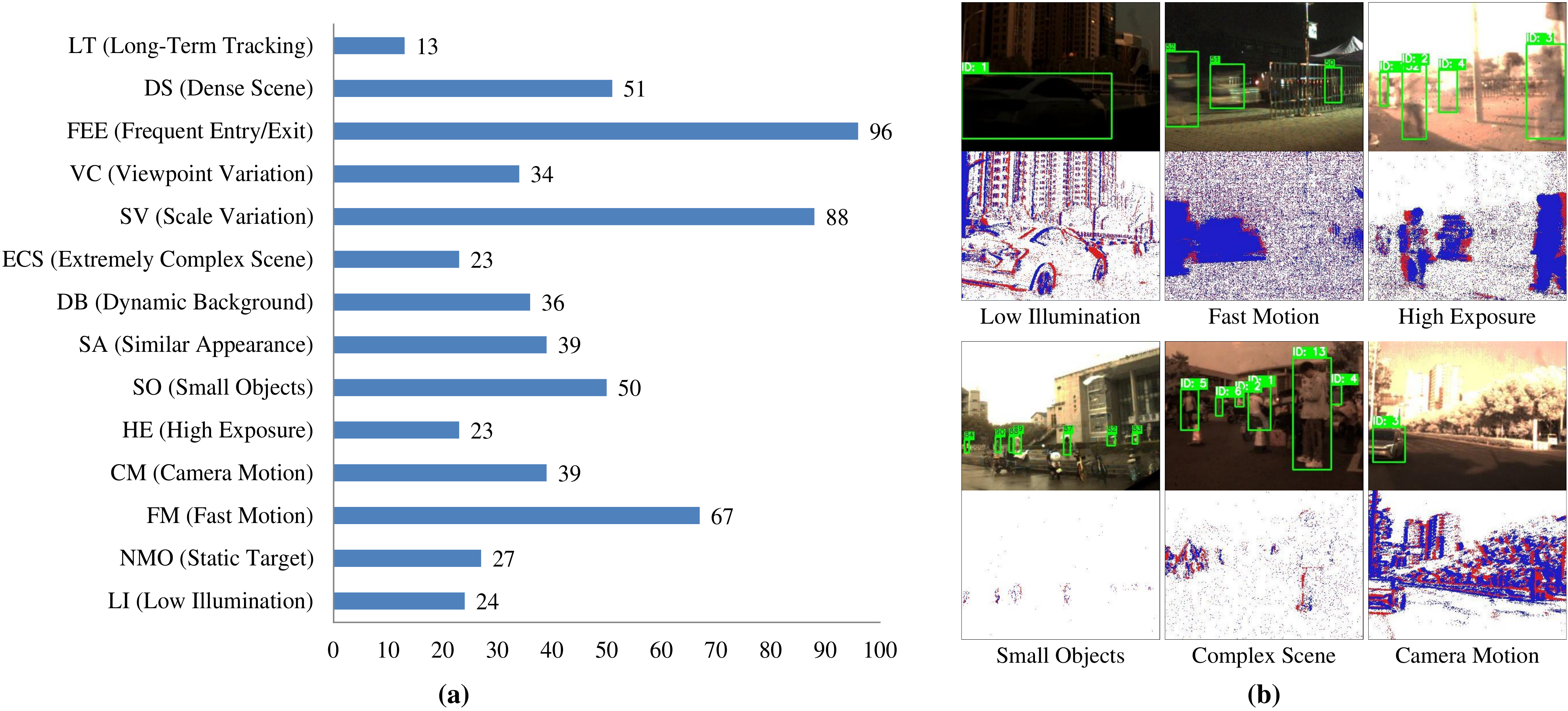}
    \caption{(a) Statistics of the 14 challenging attributes in the FEMOT dataset; (b) Representative examples from different scenes in the FEMOT dataset.}
    \label{fig:dataset_examples}
\end{figure*}

The ground-truth annotations are stored as TXT files in sequence-specific subdirectories following the MOT-style annotation format. Each annotation line contains nine standardized fields: frame ID, track ID, bounding box coordinates $(x,y)$, bounding box size $(w,h)$, validity flag (1/0), object category, and visibility score. The validity flag indicates whether the corresponding annotation is valid, while the object category is defined as 1 for pedestrians and 3 for vehicles. The visibility score provides a quantitative description of object observability, which is derived from the confidence information provided by the annotation software.

Following the file format and data organization of MOT20, we developed a standardized data format for FEMOT. Maintaining a consistent format across unimodal and multimodal datasets enables more convenient data loading and more efficient evaluation of tracking performance. Specifically, all video sequences are organized in a dedicated directory, where each sequence folder is named according to the corresponding video sequence. To distinguish modality-specific image sequences, two subdirectories are created within each sequence folder, namely \texttt{seqName\_aps} and \texttt{seqName\_dvs}. The RGB and event frames are stored in JPEG format and named according to their frame indices using a zero-padded five-digit format, e.g., \texttt{00001.jpg}, \texttt{00002.jpg}, and so on.
To further improve dataset accessibility and usability, each sequence is accompanied by a configuration file, \texttt{seqinfo.ini}, which records key metadata, including the sequence identifier, image resolution, image directory, and sequence duration. This standardized configuration allows researchers to quickly understand the structural organization and technical specifications of FEMOT, thereby facilitating dataset usage and tracker evaluation.



\begin{figure*}
\centering
\includegraphics[width=1\linewidth]{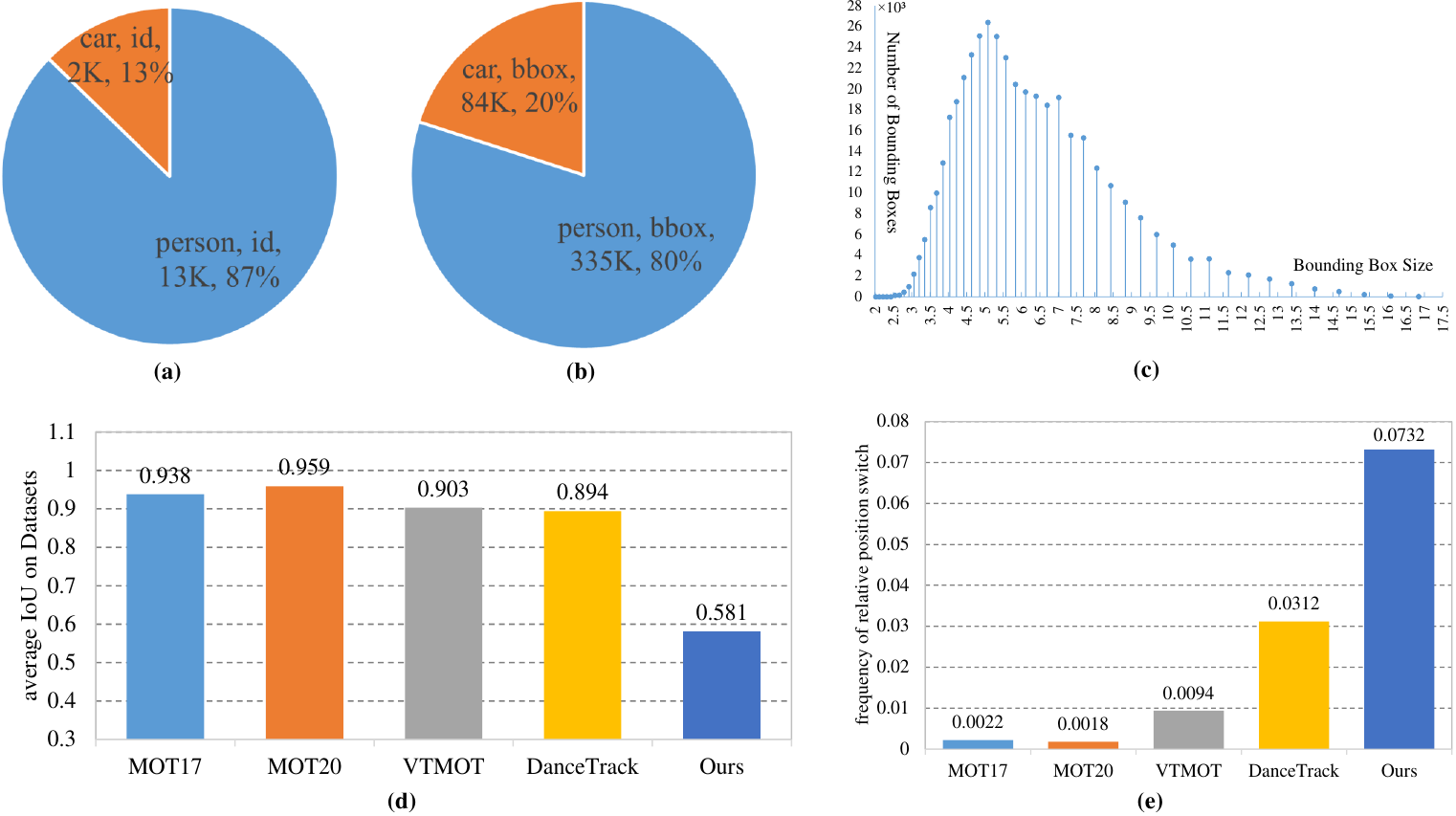}
\caption{
(a, b) The number and percentage of IDs and boxes for each category;
(c) The scale distribution of bounding boxes in our dataset;
(d) Average IoU across datasets;
(e) Frequency of relative position switches.}
\label{fig:benchmarkVIS}
\end{figure*}

\subsection{Statistical Analysis}  

\noindent \textbf{$\bullet$~14 challenging attributes.~}
As shown in Fig.~\ref{fig:dataset_examples} (a), the proposed dataset exhibits diverse and challenging attributes. Among the 14 annotated attributes, Frequent Entry/Exit (FEE) occurs most frequently, with 96 occurrences, indicating that targets often enter and leave the field of view. This characteristic poses significant challenges for track initialization, data association, and track termination. Scale Variation (SV) is also highly prominent, with 88 occurrences, reflecting substantial changes in target sizes across frames. In addition, Fast Motion (FM), Dense Scene (DS), and Small Objects (SO) are frequently observed, with 67, 51, and 50 occurrences, respectively, further increasing the difficulty of reliable detection and robust association. Overall, the attribute statistics show that the proposed dataset covers a broad spectrum of real-world tracking challenges, highlighting the complexity and diversity of the collected scenarios. These characteristics make it a comprehensive and challenging benchmark for evaluating the robustness of RGB-event multi-object tracking algorithms.

\noindent \textbf{$\bullet$~Target size and distribution.~} As shown in Fig.~\ref{fig:benchmarkVIS} (a) and Fig.~\ref{fig:benchmarkVIS} (b), our dataset mainly focuses on everyday objects, particularly people and vehicles, which commonly appear in daily scenes and are central to MOT tasks. These categories frequently occur in challenging real-world scenarios, making the dataset suitable for evaluating and improving the practical performance of MOT algorithms. As illustrated in Fig.~\ref{fig:benchmarkVIS} (c), the target-size distribution exhibits a clear long-tailed pattern, with a high proportion of small targets and progressively fewer instances as the target size increases. This indicates a scale imbalance in the dataset, where small-sized targets dominate the distribution. Such a pattern highlights the need for more robust and scale-adaptive tracking methods, especially for tiny and small objects that are frequently encountered in real-world applications.

\begin{figure}
\centering
\includegraphics[width=0.95\linewidth]{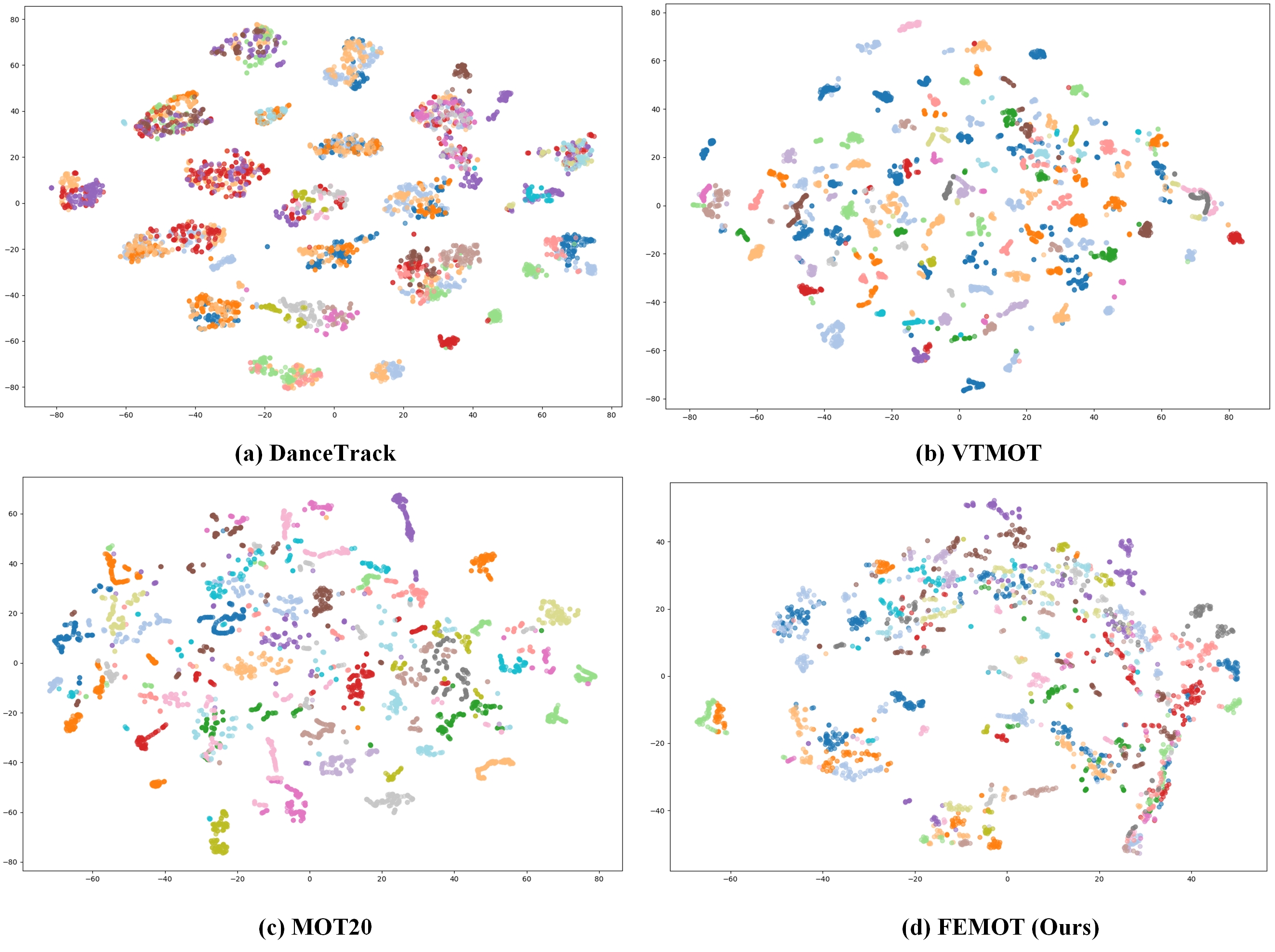}
\caption{
Visualization of re-ID features from sampled videos in
MOT20, VTMOT, DanceTrack, and FEMOT datasets using t-SNE~\cite{van2008visualizing}.}
\label{fig:tSNE}
\end{figure}

\noindent \textbf{$\bullet$~Fast Motion and Hard-to-Distinguish Appearance.~}
As shown in Fig.~\ref{fig:benchmarkVIS} (d), FEMOT yields the lowest adjacent-frame bounding-box IoU among the compared MOT datasets, suggesting substantial object displacement between consecutive frames. Such a low IoU is primarily attributed to the rapid and highly nonlinear motion patterns in FEMOT.
In addition, as illustrated in Fig.~\ref{fig:benchmarkVIS} (e), FEMOT achieves the highest Frequency of Relative Position Switches (FRPS), suggesting frequent changes in the relative spatial ordering among targets. Such frequent position switches, together with visually similar appearances, substantially increase the difficulty of identity association and make FEMOT particularly challenging for multi-object tracking.

Object appearance is a crucial cue for MOT trackers, which often rely on visual representations for target association. In MOT17 and MOT20, pedestrians exhibit noticeable variations in body scale and clothing, providing discriminative visual cues for identity association. In contrast, targets in DanceTrack often wear highly similar outfits, making appearance-based discrimination less reliable. Therefore, DanceTrack places greater emphasis on motion cues for tracking.
The VTMOT dataset contains diverse real-world scenarios in which targets usually present clear appearance differences. By contrast, FEMOT introduces additional challenges due to its varying environmental conditions. Although targets can be relatively distinguishable under well-lit conditions, low-light scenarios significantly degrade visual clarity and make accurate identification more difficult. Such illumination variability increases the complexity of tracking in FEMOT compared with existing MOT datasets.

To further illustrate these differences, we present t-SNE~\cite{van2008visualizing} visualizations of re-identification (re-ID) features sampled from MOT20, DanceTrack, VTMOT, and FEMOT in Fig.~\ref{fig:tSNE}. The visualizations show that the re-ID features in FEMOT are less separable and more entangled than those in the other datasets, indicating greater difficulty in learning discriminative appearance representations. Overall, FEMOT is designed to encourage the development of trackers that learn more robust and discriminative visual representations under challenging real-world conditions.

\subsection{Benchmarked Trackers} 
To build a comprehensive benchmark for RGB-event-based multi-object tracking, we consider a wide range of representative and state-of-the-art visual trackers from three categories: \textbf{1) tracking-by-detection trackers}, including ByteTrack~\cite{zhang2022bytetrack}, OC-SORT~\cite{cao2023observation}, MixSort-Oc~\cite{cui2023sportsmot}, MixSort-Byte~\cite{cui2023sportsmot}, Hybrid-SORT~\cite{yang2024hybrid}, SRTrack~\cite{li2024sampling}, PFTrack~\cite{zhu2025visible}, and TOPICTrack~\cite{cao2025topic}; \textbf{2) joint detection-and-tracking trackers}, including FairMOT~\cite{zhang2021fairmot}, TraDes~\cite{wu2021track}, TLDMOT~\cite{chen2024delving}, and Samba~\cite{ren2024samba}; and \textbf{3) query-based Transformer trackers}, including MOTRv2~\cite{zhang2023motrv2}, MeMOTR~\cite{gao2023memotr}, and MOTIP~\cite{gao2025multiple}. 
For a fair comparison, all trackers are retrained on the training subset of the proposed FEMOT dataset using their original training protocols, including the number of epochs and learning-rate schedules, rather than directly evaluating pretrained models on the testing subset. Due to the limited availability of multi-modal MOT algorithms, most baseline methods are originally developed for RGB-based tracking. To accommodate RGB-event inputs, we extend these methods to multi-modal variants by incorporating feature-level fusion between the RGB and event modalities. These retrained trackers serve as strong baselines, facilitating future research and benchmarking on RGB-event-based MOT.



\section{Experiments} 

\subsection{Dataset and Evaluation Metrics} 

In addition to evaluating our method on the newly proposed \textbf{FEMOT} dataset, we further compare it with state-of-the-art visual trackers on existing event-based MOT benchmarks, e.g., \textbf{DSEC-MOT dataset}~\cite{wang2024spikemot}. DSEC-MOT is a large-scale event-based MOT benchmark built upon the DSEC driving dataset, providing fine-grained annotations for real-world traffic scenarios. It contains various challenging cases, including severe occlusions, frequent trajectory intersections, and long-term re-identification, making it suitable for evaluating the robustness of MOT algorithms in complex driving scenes.

To evaluate the performance of MOT algorithms on the proposed \textbf{FEMOT} dataset, we adopt a comprehensive set of standard MOT metrics, including \textbf{Multi-Object Tracking Accuracy (MOTA)}~\cite{bernardin2008evaluating}, \textbf{Higher Order Tracking Accuracy (HOTA)}~\cite{luiten2021hota}, \textbf{IDF1}~\cite{ristani2016performance}, \textbf{Detection Accuracy (DetA)}, \textbf{Association Accuracy (AssA)}, and \textbf{Multi-Object Tracking Precision (MOTP)}. MOTA is a widely used metric that summarizes tracking errors in terms of false positives, false negatives, and identity switches; however, it does not explicitly disentangle detection quality from association quality. To provide a more fine-grained evaluation, we further report HOTA, which jointly measures tracking performance by decomposing it into DetA and AssA. In addition, IDF1 is used to evaluate identity preservation, while MOTP measures localization precision during tracking. All evaluations are conducted using TrackEval as the primary evaluation toolkit.
Since the default TrackEval protocol evaluates multi-class tracking by collapsing all categories into a single class, we additionally implement a customized script for class-aware evaluation.

The formal definitions of DetA and AssA are given by
\begin{equation}
\text{DetA} = \frac{\text{TP}*\text{Det}}{\text{TP}*\text{Det} + \text{FN}*\text{Det} + \text{FP}*\text{Det}},
\label{eq:deta}
\end{equation}
and
\begin{equation}
\text{AssA} = \frac{\text{TP}*\text{Ass}}{\text{TP}*\text{Ass} + \text{FN}*\text{Ass} + \text{FP}*\text{Ass}},
\label{eq:assa}
\end{equation}
where $\text{TP}*\text{Det}$, $\text{FN}*\text{Det}$, and $\text{FP}*\text{Det}$ denote the true positives, false negatives, and false positives for detection, respectively. Similarly, $\text{TP}*\text{Ass}$, $\text{FN}*\text{Ass}$, and $\text{FP}*\text{Ass}$ denote the corresponding quantities for data association.

\subsection{Implementation Details} 
Our tracker is built upon the MeMOTR framework~\cite{gao2023memotr} and initialized with pretrained weights of DAB-Deformable DETR~\cite{liu2022dabdetr}. Standard data augmentation strategies, including random resizing and random cropping, are applied during training. The model adopts a Transformer architecture comprising 6 encoder layers and 6 decoder layers, where the decoder is composed of 3 detection decoder layers and 3 joint Transformer decoder layers. The model is trained for 13 epochs with a batch size of 1. We adopt a progressive clip-sampling strategy, where the clip length is initially set to 2 and then increased to 3 and 4 at epochs 6 and 10, respectively. The sampling interval is set to 4. The model is optimized using AdamW~\cite{loshchilov2018adamw}, with an initial learning rate of $2\times10^{-4}$ and a weight decay of $10^{-4}$. A multi-step learning-rate scheduler is employed to decay the learning rate by a factor of 0.1 at epoch 12.
Our tracker is developed in Python based on PyTorch~\cite{paszke2019pytorch}. All experiments are conducted on a server equipped with NVIDIA A800-SXM4-80GB GPUs. More details can be found in our source code.

\subsection{Comparison on Public Benchmarks} 

\noindent \textbf{$\bullet$~Results on FEMOT Dataset.~}
As shown in Table~\ref{tab:4-4-FEMOT_fusion_table}, we conduct a comprehensive comparison with 15 state-of-the-art methods on the newly proposed FEMOT dataset. Our method consistently outperforms existing trackers in terms of HOTA, DetA, AssA, MOTA, and IDF1. Among the competing methods, MeMOTR~\cite{gao2023memotr} achieves the best baseline performance, benefiting from its temporal interaction module. However, our method further improves HOTA from 44.0 to 48.9, MOTA from 45.4 to 52.0, and IDF1 from 54.9 to 62.0. The improvements in DetA and AssA further show that our method enhances both detection accuracy and association reliability. Notably, while several tracking-by-detection methods, such as ByteTrack~\cite{zhang2022bytetrack} and SRTrack~\cite{li2024sampling}, achieve competitive detection performance, their association accuracy remains limited compared with ours. Similarly, Samba~\cite{ren2024samba} obtains strong AssA but suffers from relatively lower DetA and MOTA. In contrast, our method achieves a more balanced performance across detection, association, and identity preservation, demonstrating its effectiveness for RGB-event-based MOT on the FEMOT dataset.

\begin{table*}[!h]
    \centering
    \caption{Overall tracking performance on FEMOT dataset.} 
    \label{tab:4-4-FEMOT_fusion_table}
\resizebox{\textwidth}{!}{ 
    \begin{tabular}{l|l|c|c|cccccc} 
    \toprule
    \textbf{No.}  &\textbf{Tracker}  &\textbf{Modality} &\textbf{Source}  &\textbf{HOTA} &\textbf{DetA} 
    &\textbf{AssA} &\textbf{MOTA}     &\textbf{MOTP} &\textbf{IDF1} \\ 
    \hline 
    01  & FairMOT~\cite{zhang2021fairmot}  & RGB-Event & IJCV 2021 & 22.1 & 15.2 & 32.4 & 17.0 & \textbf{78.7} & 23.7 \\
    02  & TraDes~\cite{wu2021track} & RGB-Event & CVPR 2021 & 12.6 & 9.4 & 17.3 & 7.8 & 73.7 & 11.0 \\
    03  & ByteTrack~\cite{zhang2022bytetrack} & RGB-Event & ECCV 2022 & 36.3 & 35.4 & 38.1 & 38.0 & 73.7 & 44.3 \\
    04  & MixSort-Oc~\cite{cui2023sportsmot}   & RGB-Event & ICCV 2023 & 32.4 & 29.4 & 36.4 & 33.7 & 76.3 & 39.8 \\
    05  & MOTRv2~\cite{zhang2023motrv2} & RGB-Event & CVPR 2023 & 29.5 & 22.6 & 39.6 & 20.3 & 75.4 & 33.3 \\
    06  & OC-SORT~\cite{cao2023observation} & RGB-Event & CVPR 2023 & 23.1 & 15.0 & 36.3 & 15.9 & 77.8 & 25.0 \\
    07  & MixSort-Byte~\cite{cui2023sportsmot} & RGB-Event & ICCV 2023 & 32.1 & 36.9 & 28.6 & 36.0 & 72.9 & 37.1 \\
    08  & MeMOTR~\cite{gao2023memotr} 	& RGB-Event & ICCV 2023	& 44.0 & 40.2 & 49.0 & 45.4 & 76.9 & 54.9 \\
    09  & TLDMOT~\cite{chen2024delving} & RGB-Event & CVPR 2024 & 28.2 & 33.6 & 24.7 & 23.6 & 73.0 & 33.4 \\
    10  & Hybrid-SORT~\cite{yang2024hybrid} & RGB-Event & AAAI 2024 & 22.1 & 13.8 & 36.2 & 14.2 & 76.6 & 23.1 \\
    11  & SRTrack~\cite{li2024sampling} & RGB-Event & AAAI 2024 & 38.5 & 35.8 & 42.0 & 40.8 & 76.0 & 47.0 \\
    12  & PFTrack~\cite{zhu2025visible} & RGB-Event & PR 2025 & 29.7 & 33.5 & 27.0 & 33.6 & 75.7 & 34.0 \\
    13  & TOPICTrack~\cite{cao2025topic} & RGB-Event & TIP 2025  & 17.9 & 16.2 & 20.3 & 14.4 & 76.2 & 17.3 \\
    14  & Samba~\cite{ren2024samba} & RGB-Event & ICLR 2025  & 38.1 & 30.0 & 49.1 & 33.7 & 75.8 & 48.1 \\
    15  & MOTIP~\cite{gao2025multiple}  & RGB-Event & CVPR 2025 & 33.2 & 36.6 & 30.9 & 16.9 & 74.4 & 36.6 \\
    \hline
    16  & Ours & RGB-Event & -  & \textbf{48.9} & \textbf{45.2} & \textbf{53.8} & \textbf{52.0} & 76.7 & \textbf{62.0}  \\
    \bottomrule
    \end{tabular}
    }
\end{table*}

\begin{table*}
\centering
\caption{Overall tracking performance on DSEC-MOT dataset.} 
\label{tab:DSEC-MOT} 
\resizebox{\textwidth}{!}{ 
\begin{tabular}{l|l|c|c|ccccc} 
\toprule
\textbf{No.}  &\textbf{Trackers}  &\textbf{Modality} &\textbf{Source}  &\textbf{HOTA} &\textbf{DetA} 
&\textbf{AssA} &\textbf{MOTA}     &\textbf{IDF1} \\ 
\hline 
01 &GTR~\cite{zhang2021fairmot} &Event &CVPR 2022 &41.9 &43.8 &40.4 &47.5 &49.0 \\
02 &SiamMOT~\cite{shuai2021siammot} &Event &CVPR 2021 &49.1 &47.4 &51.4 &44.4 &59.1 \\
03 &ByteTrack~\cite{zhang2022bytetrack}   &Event &ECCV 2022 &44.9 &42.3 &47.9 &48.1 &54.5 \\
04 &Trackformer~\cite{meinhardt2022trackformer} &Event &CVPR 2022 &45.9 &47.2 &45.0 &48.6 &56.2 \\
05 &SpikeMOT~\cite{wang2024spikemot} &Event & IEEE Access 2025 &52.5 &\textbf{49.5} &55.7 &\textbf{54.7} &62.9 \\
\hline
06 &OC-SORT~\cite{cao2023observation} &RGB-Event &CVPR 2023 &27.7 &17.7 &43.9 &10.6 &29.8 \\
07 &MeMOTR~\cite{gao2023memotr} &RGB-Event &ICCV 2023 &48.4	&36.2	&64.9	&35.8	&54.9\\
08 &Samba~\cite{ren2024samba} &RGB-Event &ICLR 2025 &\textbf{55.4}	&46.4	&66.5	&50.3	&\textbf{64.9} \\
09 &MOTIP~\cite{gao2025multiple} &RGB-Event &CVPR 2025 &37.2	&32.2	&43.3	&17.8 &42.8	\\
\hline
10 &\textbf{Ours} &RGB-Event &- &52.4 &40.6 &\textbf{67.8} &37.5 &59.0 \\
\bottomrule
\end{tabular}}
\end{table*}

\noindent \textbf{$\bullet$~Results on DSEC-MOT Dataset.~} 
We also report the tracking performance on the DSEC-MOT dataset, as shown in Table~\ref{tab:DSEC-MOT}. Among the event-only methods, SpikeMOT~\cite{wang2024spikemot} achieves strong performance, obtaining the best DetA and MOTA scores. For RGB-event-based methods, Samba~\cite{ren2024samba} achieves the highest HOTA and IDF1 scores, indicating strong overall tracking accuracy and identity preservation.
Our method achieves the best AssA score of 67.8, outperforming all compared methods, including SpikeMOT and Samba. This result suggests that our tracker can effectively maintain object identities under challenging scenarios. However, our HOTA and IDF1 scores are lower than those of SpikeMOT and Samba, indicating that the detection quality still limits the overall tracking performance on this event-based benchmark. Overall, these results demonstrate the effectiveness of our method in association modeling, while suggesting that further improvements in detection accuracy could enhance its overall performance on DSEC-MOT.


\subsection{Ablation Study}

\noindent \textbf{$\bullet$~Component Analysis.~}
To demonstrate the effectiveness of each module design, we analyze the contribution of each component, as shown in Table~\ref{tab:Component_Analysis}. 1) Removing the shared multi-scale weights leads to the most significant performance degradation, with HOTA decreasing from 48.9 to 40.5 and IDF1 dropping from 62.0 to 50.9. This indicates that shared multi-scale weights are essential for learning effective multi-scale representations. 2) When the frequency-aware feature fusion module is removed, the model suffers substantial performance drops across most metrics. Specifically, HOTA decreases from 48.9 to 44.0, AssA drops from 53.8 to 49.0, and IDF1 decreases from 62.0 to 54.9. These results demonstrate that frequency-aware feature fusion is crucial for effectively integrating multi-modal information. 3) Removing the modality-specific branches decreases HOTA from 48.9 to 45.5 and IDF1 from 62.0 to 57.3, indicating that these branches play an important role in capturing modality-dependent representations. 4) Removing the amplitude/phase attention module slightly improves DetA and MOTA, but reduces HOTA, AssA, and IDF1. This suggests that amplitude/phase attention mainly benefits association quality and identity consistency. Overall, FEMOTR achieves the best HOTA, AssA, and IDF1 scores, validating the effectiveness of the proposed components and showing that the full model achieves a better balance between detection and association performance.

\begin{table}[!htp]
\centering
\caption{Component Analysis on FEMOT dataset.} 
\label{tab:Component_Analysis}
\resizebox{\textwidth}{!}{ 
\begin{tabular}{l|c|cccccc} 
\hline
\textbf{\#} &\textbf{Model Variant} & \textbf{HOTA} & \textbf{DetA} 
& \textbf{AssA} & \textbf{MOTA} & \textbf{MOTP} & \textbf{IDF1} \\ 
\hline 
1 &w/o Shared Multi-scale Weights & 40.5 & 37.1 & 45.1 & 42.8 & 76.8 & 50.9 \\
2 &w/o Frequency Aware Feature Fusion & 44.0 & 40.2 & 49.0 & 45.4 & \textbf{76.9} & 54.9 \\
3 &w/o Modality Specific Branch & 45.5 & 41.2 & 51.0 & 47.5 & 76.7 & 57.3 \\
4 &w/o Amplitude/Phase Attention & 48.3 & \textbf{45.7} & 52.0 & \textbf{52.5} & 76.6 & 61.1 \\
\hline
5 &\textbf{FEMOTR (Ours)} & \textbf{48.9} & 45.2 & \textbf{53.8} & 52.0 & 76.7 & \textbf{62.0} \\
\hline
\end{tabular}
}
\end{table}

\noindent \textbf{$\bullet$~Ablation Study of Different Fusion Strategies.~}
As shown in Table~\ref{tab:Fusion_Strategies}, we compare the proposed frequency-aware feature fusion method with several state-of-the-art fusion methods. Simple fusion strategies, such as Add and Concatenate, improve upon the baseline, indicating that RGB and event modalities provide complementary cues for MOT. However, existing frequency-based fusion methods, including FECNet, FCFE, and FMAP, fail to deliver consistent gains and even lead to performance degradation on several metrics, suggesting that generic frequency-domain fusion modules are not directly suitable for RGB-event MOT.
By contrast, our method achieves the best performance on most metrics. These results demonstrate that the proposed fusion strategy can more effectively exploit complementary multimodal cues, thereby improving both detection accuracy and association quality.

\begin{table*}
\centering
\caption{Experimental Results of Different Fusion Strategies.}  
\label{tab:Fusion_Strategies}
\small 
\resizebox{\textwidth}{!}{ 
\begin{tabular}{l|l|cccccc} 
\hline
\textbf{\#} &\textbf{Configuration} & \textbf{HOTA} & \textbf{DetA} 
& \textbf{AssA} & \textbf{MOTA} & \textbf{MOTP} & \textbf{IDF1} \\ 
\hline 
1 &Baseline & 44.0 & 40.2 & 49.0 & 45.4 & 76.9 & 54.9 \\
2 &Add & 47.1 & 44.0 & 51.3 & 50.0 & 76.7 & 59.1 \\
3 &Concatenate & 45.5 & 41.2 & 51.0 & 47.5 & 76.7 & 57.3 \\
\hline 
4 &FECNet~\cite{ren2024frequency} & 39.2 & 32.3 & 48.3 & 20.7 & 76.0 & 48.7 \\
5 &FCFE~\cite{kim2024frequency} & 41.0 & 37.6 & 45.7 & 47.5 & \textbf{77.0} & 51.0 \\
6 &FMAP~\cite{xu2025learning} & 40.8 & 36.6 & 46.4 & 41.8 & 76.7 & 51.1 \\
\hline
7 &\textbf{FAF (Ours)} & \textbf{48.9} & \textbf{45.2} & \textbf{53.8} & \textbf{52.0} & 76.7 & \textbf{62.0} \\
\hline
\end{tabular}
}
\end{table*}

\noindent \textbf{$\bullet$~Ablation Study of Different Input Modalities.~}
As shown in Table~\ref{tab:femot_modality_comparison}, we conduct an ablation study to analyze the contribution of different input modalities. For the baseline model, the dual-modal input clearly outperforms both RGB-only and event-only settings, indicating that RGB and event data provide complementary information for MOT. However, the event-only setting performs poorly in DetA and MOTA, suggesting that event data alone is insufficient for reliable object detection.
Our method achieves the best performance when both RGB and event modalities are used. Compared with the RGB-only variant, the dual-modal setting improves HOTA from 42.5 to 48.9 and IDF1 from 53.9 to 62.0. It also brings notable gains in DetA, AssA, and MOTA, demonstrating that the proposed method can effectively integrate multimodal cues to improve both detection and association performance.

\begin{table}[!htp]
\centering
\caption{Performance comparison of different modality configurations on the FEMOT dataset.}
\label{tab:femot_modality_comparison}
\resizebox{\textwidth}{!}{ 
\begin{tabular}{l|l|cccccc} 
\hline
\textbf{Methods} & \textbf{Modality} & \textbf{HOTA} & \textbf{DetA} & \textbf{AssA} & \textbf{MOTA} & \textbf{MOTP} & \textbf{IDF1} \\ 
\hline
Baseline & Event & 17.7 & 7.8 & 40.1 & 5.9 & 73.7 & 15.5 \\
Baseline & RGB & 23.7 & 16.9 & 34.2 & 16.5 & 74.6 & 27.1 \\
Baseline & RGB \& Event & 44.0 & 40.2 & 49.0 & 45.4 & \textbf{76.9} & 54.9 \\
\hline 
Ours & Event & 14.3 & 6.6 & 32.3 & 2.3 & 70.8 & 11.8 \\
Ours & RGB & 42.5 & 37.4 & 49.1 & 44.0 & 76.9 & 53.9 \\
Ours & RGB \& Event & \textbf{48.9} & \textbf{45.2} & \textbf{53.8} & \textbf{52.0} & 76.7 & \textbf{62.0} \\
\hline
\end{tabular}
}
\end{table}

\begin{figure}[!h]
    \centering
    \includegraphics[width=0.9\linewidth]{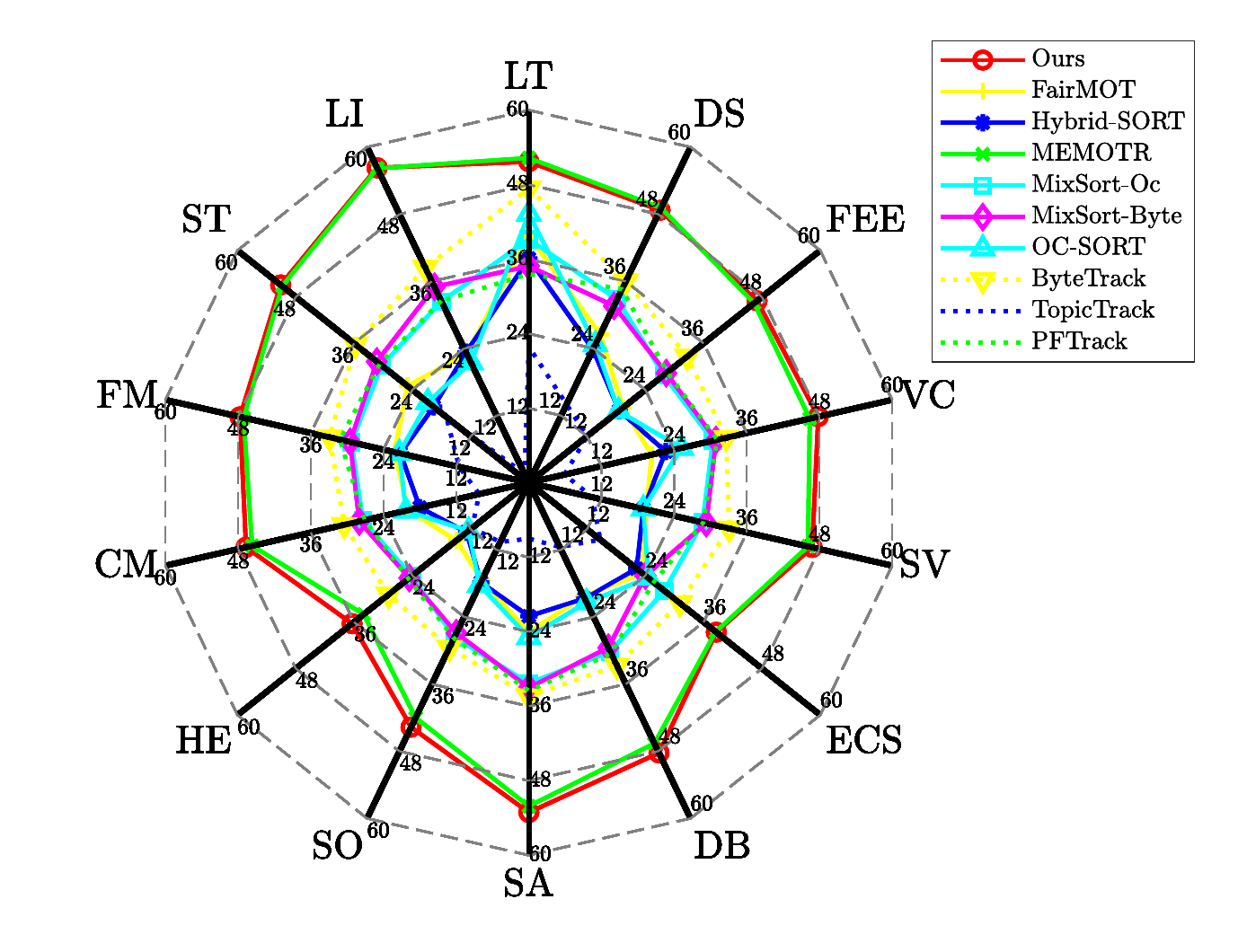}
    \caption{Tracking results under each challenging attribute.}
    \label{fig:femot_radar}
\end{figure}

\noindent \textbf{$\bullet$~Analysis under Specific Challenging Attributes.~}
Our newly proposed dataset covers 14 real-world challenging attributes. As shown in Fig.~\ref{fig:femot_radar}, we compare the proposed method with state-of-the-art trackers under each attribute. The proposed method achieves the best or highly competitive performance in most challenging scenarios, indicating its strong robustness to complex tracking conditions. In particular, it shows clear advantages under challenging attributes such as HE (High Exposure), SO (Small Objects), SA (Similar Appearance), and DB (Dynamic Background), where accurate object detection and reliable identity association are difficult.
The results suggest that our method effectively exploits the complementary properties of RGB and event modalities. RGB frames provide rich appearance and semantic information, whereas event data captures high-frequency contour cues. By integrating these complementary cues, the proposed method achieves more stable tracking performance across diverse challenging environments than existing trackers.

\subsection{Visualization}

\begin{figure*}
    \centering
    \includegraphics[width=0.98\linewidth]{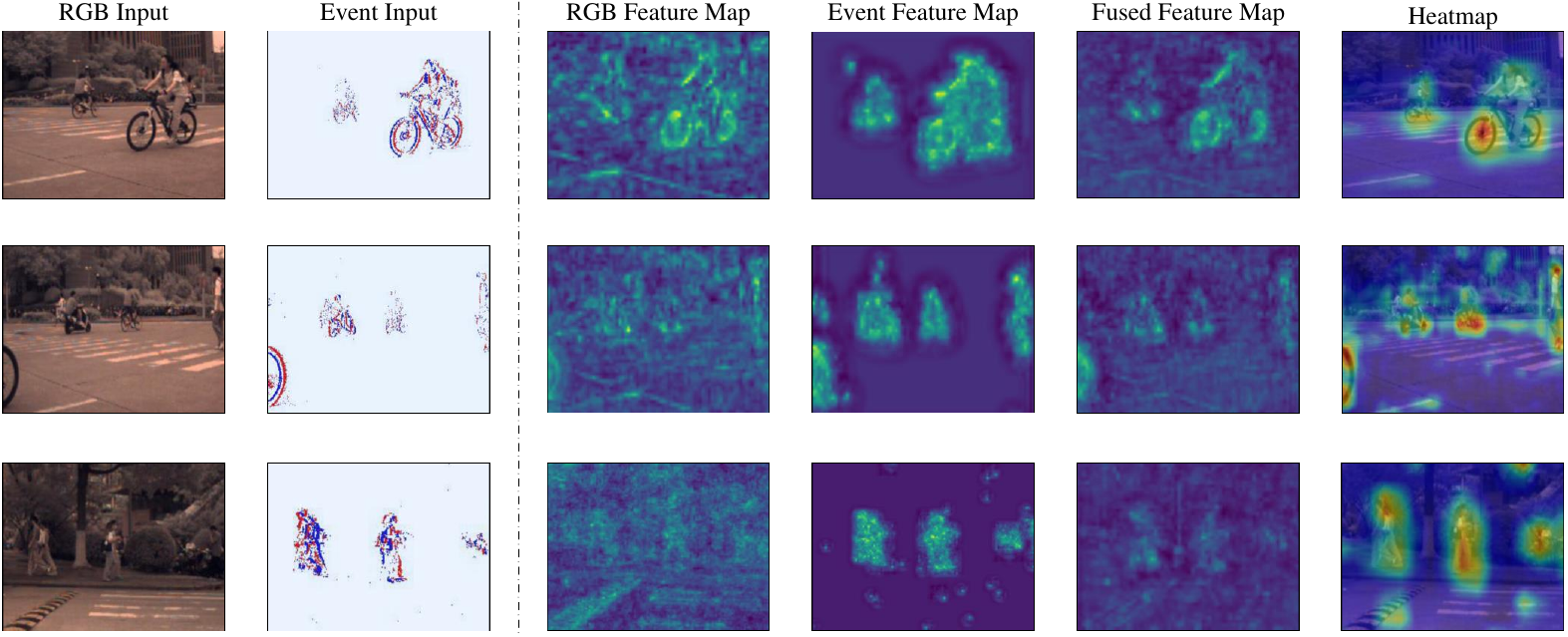}
    \caption{Visualization of feature maps before and after frequency fusion, with corresponding heatmap overlays on RGB frames.}
    \label{fig:feature_heatmap}
\end{figure*}

\noindent \textbf{$\bullet$~Visualization of Feature Maps and Heatmaps.~}
In addition to the quantitative analysis presented above, we further provide visual analyses of the proposed tracking algorithm to offer a more intuitive understanding of our tracking framework. As shown in Fig.~\ref{fig:feature_heatmap}, we visualize the feature maps before and after frequency-aware fusion, together with the corresponding heatmap overlays on RGB frames. The frequency-aware fusion module enables the fused feature maps to generate more compact and target-aware activations while effectively suppressing irrelevant background responses.
In the heatmap overlays, brighter regions indicate areas that receive higher attention from the model. The results show that our model can accurately focus on object regions in complex scenes. These visualizations demonstrate that our frequency-aware fusion strategy effectively enhances multimodal feature representation, thereby improving target localization and tracking robustness in RGB-event MOT.

\begin{figure}
\centering
\includegraphics[width=0.9\linewidth]{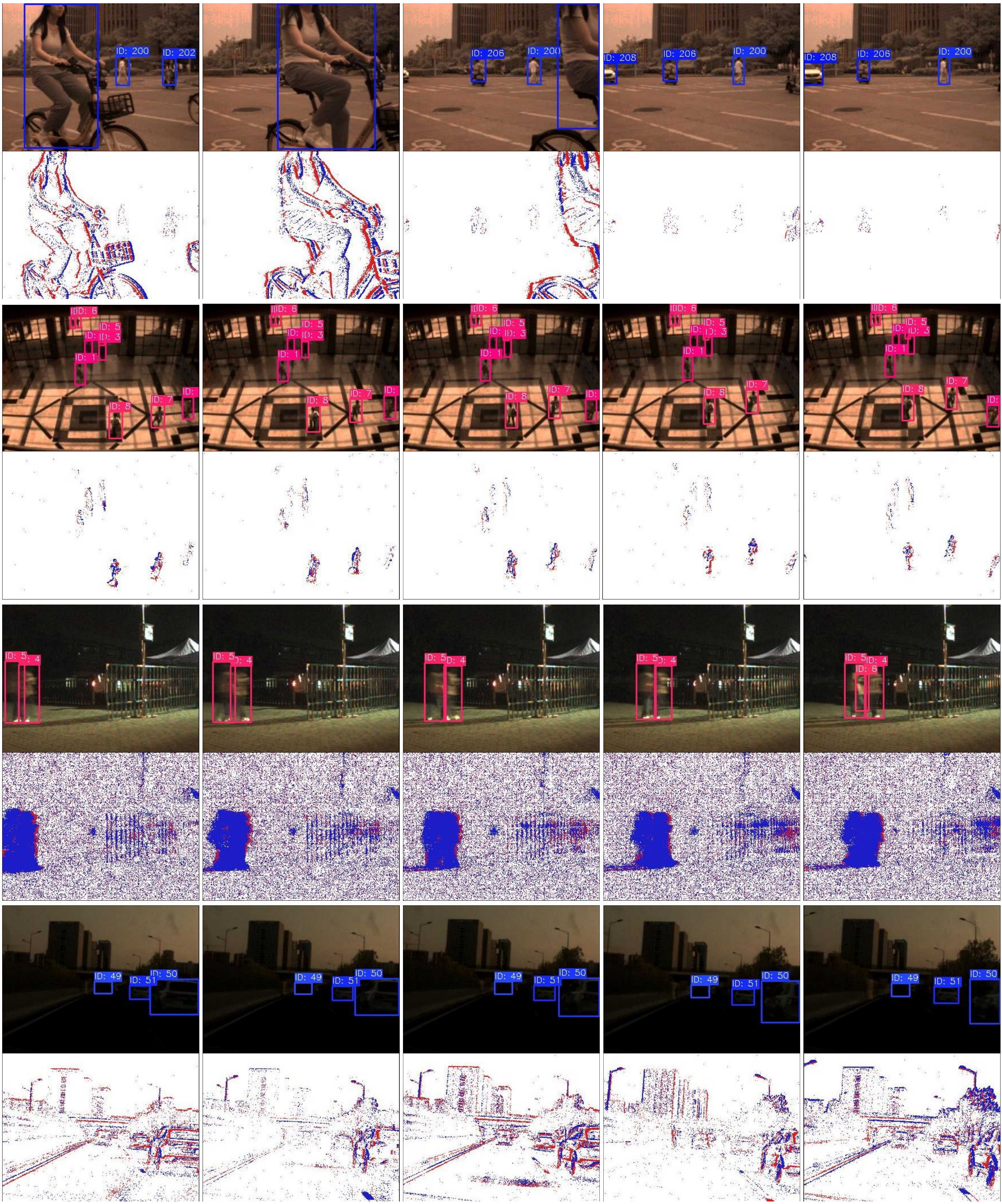}
\caption{Tracking results produced by our method on the FEMOT dataset.}
\label{fig:tracking_result}
\end{figure}

\noindent \textbf{$\bullet$~Visualization of tracking results.~}
Fig.~\ref{fig:tracking_result} visualizes the tracking results produced by our method on the FEMOT dataset. Each scene is presented with the RGB frame and its corresponding event representation. The results demonstrate that our method maintains temporally consistent identities across diverse and challenging scenarios. Specifically, in the daytime bicycle scene, the tracker accurately follows large foreground objects while preserving the identities of small distant targets. In the top-down view, multiple objects with similar appearances are tracked stably despite dense spatial layouts and shadow interference. In the nighttime sequence, where RGB observations are severely degraded by motion blur and low illumination, the event stream still provides clear motion cues, enabling robust localization and reliable identity maintenance.
Overall, these visualizations confirm that the proposed method effectively exploits event-based motion information to complement RGB appearance cues, thereby further validating its robustness under challenging real-world tracking conditions.

\subsection{Limitations and Future Work}  
Although the proposed method achieves robust tracking performance by exploiting complementary RGB and event information, there remains room for further improvement. First, the current framework does not dynamically adjust the event density according to scene variations. Fast-moving targets usually generate dense event streams, whereas slow-moving targets produce relatively sparse event responses. Therefore, dynamically adapting the event accumulation window to different motion speeds could help construct more informative and balanced event representations, which remains a promising direction for future improvement.

Second, the natural language modality is not considered in the current framework. Language descriptions could provide high-level semantic priors, such as object categories, spatial relations, and motion intentions, which may further improve the accuracy of target localization and identity association. Incorporating language-guided cues into RGB-event tracking could therefore be an interesting direction for future work.

\section{Conclusion}  
In this paper, we introduce FEMOT, a new large-scale RGB-event multi-object tracking dataset that bridges the data gap in this field. FEMOT covers 14 challenging real-world attributes, capturing the complexity and diversity of practical scenarios. Meanwhile, over ten recent strong trackers are retrained and evaluated on FEMOT, establishing it as a comprehensive benchmark for future comparisons.
Built upon FEMOT, we further propose FEMOTR, a novel multimodal tracking framework for RGB-event multi-object tracking. FEMOTR effectively exploits low-frequency appearance and semantic cues from RGB frames and high-frequency contour cues from event data. By integrating multimodal features in the frequency domain, it facilitates robust object localization and identity association.

\section*{Acknowledgment}
This work was supported by the National Natural Science Foundation of China under Grant 62102205. Anhui Provincial Natural Science Foundation-Outstanding Youth Project, 2408085Y032. The authors acknowledge the High-performance Computing Platform of Anhui University for providing computing resources.

\section*{Data Availability Statement} 
\noindent The authors confirm that the data supporting this work's findings are available within the article or its supplementary materials.


\bibliography{reference}


\begin{thebibliography}{62}
\ifx \bisbn   \undefined \def \bisbn  #1{ISBN #1}\fi
\ifx \binits  \undefined \def \binits#1{#1}\fi
\ifx \bauthor  \undefined \def \bauthor#1{#1}\fi
\ifx \batitle  \undefined \def \batitle#1{#1}\fi
\ifx \bjtitle  \undefined \def \bjtitle#1{#1}\fi
\ifx \bvolume  \undefined \def \bvolume#1{\textbf{#1}}\fi
\ifx \byear  \undefined \def \byear#1{#1}\fi
\ifx \bissue  \undefined \def \bissue#1{#1}\fi
\ifx \bfpage  \undefined \def \bfpage#1{#1}\fi
\ifx \blpage  \undefined \def \blpage #1{#1}\fi
\ifx \burl  \undefined \def \burl#1{\textsf{#1}}\fi
\ifx \doiurl  \undefined \def \doiurl#1{\url{https://doi.org/#1}}\fi
\ifx \betal  \undefined \def \betal{\textit{et al.}}\fi
\ifx \binstitute  \undefined \def \binstitute#1{#1}\fi
\ifx \binstitutionaled  \undefined \def \binstitutionaled#1{#1}\fi
\ifx \bctitle  \undefined \def \bctitle#1{#1}\fi
\ifx \beditor  \undefined \def \beditor#1{#1}\fi
\ifx \bpublisher  \undefined \def \bpublisher#1{#1}\fi
\ifx \bbtitle  \undefined \def \bbtitle#1{#1}\fi
\ifx \bedition  \undefined \def \bedition#1{#1}\fi
\ifx \bseriesno  \undefined \def \bseriesno#1{#1}\fi
\ifx \blocation  \undefined \def \blocation#1{#1}\fi
\ifx \bsertitle  \undefined \def \bsertitle#1{#1}\fi
\ifx \bsnm \undefined \def \bsnm#1{#1}\fi
\ifx \bsuffix \undefined \def \bsuffix#1{#1}\fi
\ifx \bparticle \undefined \def \bparticle#1{#1}\fi
\ifx \barticle \undefined \def \barticle#1{#1}\fi
\bibcommenthead
\ifx \bconfdate \undefined \def \bconfdate #1{#1}\fi
\ifx \botherref \undefined \def \botherref #1{#1}\fi
\ifx \url \undefined \def \url#1{\textsf{#1}}\fi
\ifx \bchapter \undefined \def \bchapter#1{#1}\fi
\ifx \bbook \undefined \def \bbook#1{#1}\fi
\ifx \bcomment \undefined \def \bcomment#1{#1}\fi
\ifx \oauthor \undefined \def \oauthor#1{#1}\fi
\ifx \citeauthoryear \undefined \def \citeauthoryear#1{#1}\fi
\ifx \endbibitem  \undefined \def \endbibitem {}\fi
\ifx \bconflocation  \undefined \def \bconflocation#1{#1}\fi
\ifx \arxivurl  \undefined \def \arxivurl#1{\textsf{#1}}\fi
\csname PreBibitemsHook\endcsname

\bibitem{zhang2021fairmot}
\begin{barticle}
\bauthor{\bsnm{Zhang}, \binits{Y.}},
\bauthor{\bsnm{Wang}, \binits{C.}},
\bauthor{\bsnm{Wang}, \binits{X.}},
\bauthor{\bsnm{Zeng}, \binits{W.}},
\bauthor{\bsnm{Liu}, \binits{W.}}:
\batitle{Fairmot: On the fairness of detection and re-identification in
  multiple object tracking}.
\bjtitle{International journal of computer vision}
\bvolume{129},
\bfpage{3069}--\blpage{3087}
(\byear{2021})
\end{barticle}
\endbibitem

\bibitem{zhang2022bytetrack}
\begin{bchapter}
\bauthor{\bsnm{Zhang}, \binits{Y.}},
\bauthor{\bsnm{Sun}, \binits{P.}},
\bauthor{\bsnm{Jiang}, \binits{Y.}},
\bauthor{\bsnm{Yu}, \binits{D.}},
\bauthor{\bsnm{Weng}, \binits{F.}},
\bauthor{\bsnm{Yuan}, \binits{Z.}},
\bauthor{\bsnm{Luo}, \binits{P.}},
\bauthor{\bsnm{Liu}, \binits{W.}},
\bauthor{\bsnm{Wang}, \binits{X.}}:
\bctitle{Bytetrack: Multi-object tracking by associating every detection box}.
In: \bbtitle{European Conference on Computer Vision},
pp. \bfpage{1}--\blpage{21}
(\byear{2022}).
\bcomment{Springer}
\end{bchapter}
\endbibitem

\bibitem{gao2023memotr}
\begin{bchapter}
\bauthor{\bsnm{Gao}, \binits{R.}},
\bauthor{\bsnm{Wang}, \binits{L.}}:
\bctitle{Memotr: Long-term memory-augmented transformer for multi-object
  tracking}.
In: \bbtitle{Proceedings of the IEEE/CVF International Conference on Computer
  Vision},
pp. \bfpage{9901}--\blpage{9910}
(\byear{2023})
\end{bchapter}
\endbibitem

\bibitem{gao2025multiple}
\begin{bchapter}
\bauthor{\bsnm{Gao}, \binits{R.}},
\bauthor{\bsnm{Qi}, \binits{J.}},
\bauthor{\bsnm{Wang}, \binits{L.}}:
\bctitle{Multiple object tracking as id prediction}.
In: \bbtitle{Proceedings of the Computer Vision and Pattern Recognition
  Conference},
pp. \bfpage{27883}--\blpage{27893}
(\byear{2025})
\end{bchapter}
\endbibitem

\bibitem{ren2024samba}
\begin{botherref}
\oauthor{\bsnm{Ren}, \binits{L.}},
\oauthor{\bsnm{Liu}, \binits{Y.}},
\oauthor{\bsnm{Lu}, \binits{Y.}},
\oauthor{\bsnm{Shen}, \binits{Y.}},
\oauthor{\bsnm{Liang}, \binits{C.}},
\oauthor{\bsnm{Chen}, \binits{W.}}:
Samba: Simple hybrid state space models for efficient unlimited context
  language modeling.
arXiv preprint arXiv:2406.07522
(2024)
\end{botherref}
\endbibitem

\bibitem{cao2025topic}
\begin{barticle}
\bauthor{\bsnm{Cao}, \binits{X.}},
\bauthor{\bsnm{Zheng}, \binits{Y.}},
\bauthor{\bsnm{Yao}, \binits{Y.}},
\bauthor{\bsnm{Qin}, \binits{H.}},
\bauthor{\bsnm{Cao}, \binits{X.}},
\bauthor{\bsnm{Guo}, \binits{S.}}:
\batitle{Topic: A parallel association paradigm for multi-object tracking under
  complex motions and diverse scenes}.
\bjtitle{IEEE Transactions on Image Processing}
\bvolume{34},
\bfpage{743}--\blpage{758}
(\byear{2025})
\end{barticle}
\endbibitem

\bibitem{milan2016mot16}
\begin{botherref}
\oauthor{\bsnm{Milan}, \binits{A.}},
\oauthor{\bsnm{Leal-Taix{\'e}}, \binits{L.}},
\oauthor{\bsnm{Reid}, \binits{I.}},
\oauthor{\bsnm{Roth}, \binits{S.}},
\oauthor{\bsnm{Schindler}, \binits{K.}}:
Mot16: A benchmark for multi-object tracking.
arXiv preprint arXiv:1603.00831
(2016)
\end{botherref}
\endbibitem

\bibitem{dendorfer2020mot20}
\begin{botherref}
\oauthor{\bsnm{Dendorfer}, \binits{P.}},
\oauthor{\bsnm{Rezatofighi}, \binits{H.}},
\oauthor{\bsnm{Milan}, \binits{A.}},
\oauthor{\bsnm{Shi}, \binits{J.}},
\oauthor{\bsnm{Cremers}, \binits{D.}},
\oauthor{\bsnm{Reid}, \binits{I.}},
\oauthor{\bsnm{Roth}, \binits{S.}},
\oauthor{\bsnm{Schindler}, \binits{K.}},
\oauthor{\bsnm{Leal-Taix{\'e}}, \binits{L.}}:
Mot20: A benchmark for multi object tracking in crowded scenes.
arXiv preprint arXiv:2003.09003
(2020)
\end{botherref}
\endbibitem

\bibitem{sun2022dancetrack}
\begin{bchapter}
\bauthor{\bsnm{Sun}, \binits{P.}},
\bauthor{\bsnm{Cao}, \binits{J.}},
\bauthor{\bsnm{Jiang}, \binits{Y.}},
\bauthor{\bsnm{Yuan}, \binits{Z.}},
\bauthor{\bsnm{Bai}, \binits{S.}},
\bauthor{\bsnm{Kitani}, \binits{K.}},
\bauthor{\bsnm{Luo}, \binits{P.}}:
\bctitle{Dancetrack: Multi-object tracking in uniform appearance and diverse
  motion}.
In: \bbtitle{Proceedings of the IEEE/CVF Conference on Computer Vision and
  Pattern Recognition},
pp. \bfpage{20993}--\blpage{21002}
(\byear{2022})
\end{bchapter}
\endbibitem

\bibitem{cui2023sportsmot}
\begin{bchapter}
\bauthor{\bsnm{Cui}, \binits{Y.}},
\bauthor{\bsnm{Zeng}, \binits{C.}},
\bauthor{\bsnm{Zhao}, \binits{X.}},
\bauthor{\bsnm{Yang}, \binits{Y.}},
\bauthor{\bsnm{Wu}, \binits{G.}},
\bauthor{\bsnm{Wang}, \binits{L.}}:
\bctitle{Sportsmot: A large multi-object tracking dataset in multiple sports
  scenes}.
In: \bbtitle{Proceedings of the IEEE/CVF International Conference on Computer
  Vision},
pp. \bfpage{9921}--\blpage{9931}
(\byear{2023})
\end{bchapter}
\endbibitem

\bibitem{dave2020tao}
\begin{bchapter}
\bauthor{\bsnm{Dave}, \binits{A.}},
\bauthor{\bsnm{Khurana}, \binits{T.}},
\bauthor{\bsnm{Tokmakov}, \binits{P.}},
\bauthor{\bsnm{Schmid}, \binits{C.}},
\bauthor{\bsnm{Ramanan}, \binits{D.}}:
\bctitle{Tao: A large-scale benchmark for tracking any object}.
In: \bbtitle{Computer Vision--ECCV 2020: 16th European Conference, Glasgow, UK,
  August 23--28, 2020, Proceedings, Part V 16},
pp. \bfpage{436}--\blpage{454}
(\byear{2020}).
\bcomment{Springer}
\end{bchapter}
\endbibitem

\bibitem{yu2020bdd100k}
\begin{bchapter}
\bauthor{\bsnm{Yu}, \binits{F.}},
\bauthor{\bsnm{Chen}, \binits{H.}},
\bauthor{\bsnm{Wang}, \binits{X.}},
\bauthor{\bsnm{Xian}, \binits{W.}},
\bauthor{\bsnm{Chen}, \binits{Y.}},
\bauthor{\bsnm{Liu}, \binits{F.}},
\bauthor{\bsnm{Madhavan}, \binits{V.}},
\bauthor{\bsnm{Darrell}, \binits{T.}}:
\bctitle{Bdd100k: A diverse driving dataset for heterogeneous multitask
  learning}.
In: \bbtitle{Proceedings of the IEEE/CVF Conference on Computer Vision and
  Pattern Recognition},
pp. \bfpage{2636}--\blpage{2645}
(\byear{2020})
\end{bchapter}
\endbibitem

\bibitem{gallego2020event}
\begin{barticle}
\bauthor{\bsnm{Gallego}, \binits{G.}},
\bauthor{\bsnm{Delbr{\"u}ck}, \binits{T.}},
\bauthor{\bsnm{Orchard}, \binits{G.}},
\bauthor{\bsnm{Bartolozzi}, \binits{C.}},
\bauthor{\bsnm{Taba}, \binits{B.}},
\bauthor{\bsnm{Censi}, \binits{A.}},
\bauthor{\bsnm{Leutenegger}, \binits{S.}},
\bauthor{\bsnm{Davison}, \binits{A.J.}},
\bauthor{\bsnm{Conradt}, \binits{J.}},
\bauthor{\bsnm{Daniilidis}, \binits{K.}}, \betal:
\batitle{Event-based vision: A survey}.
\bjtitle{IEEE transactions on pattern analysis and machine intelligence}
\bvolume{44}(\bissue{1}),
\bfpage{154}--\blpage{180}
(\byear{2020})
\end{barticle}
\endbibitem

\bibitem{wang2024spikemot}
\begin{barticle}
\bauthor{\bsnm{Wang}, \binits{S.}},
\bauthor{\bsnm{Wang}, \binits{Z.}},
\bauthor{\bsnm{Li}, \binits{C.}},
\bauthor{\bsnm{Qi}, \binits{X.}},
\bauthor{\bsnm{So}, \binits{H.K.-H.}}:
\batitle{Spikemot: Event-based multi-object tracking with sparse motion
  features}.
\bjtitle{IEEE Access}
\bvolume{13},
\bfpage{214}--\blpage{230}
(\byear{2024})
\end{barticle}
\endbibitem

\bibitem{cui2026peod}
\begin{barticle}
\bauthor{\bsnm{Cui}, \binits{L.}},
\bauthor{\bsnm{Liu}, \binits{H.}},
\bauthor{\bsnm{Liu}, \binits{M.}},
\bauthor{\bsnm{Lin}, \binits{E.}},
\bauthor{\bsnm{Jiang}, \binits{D.}},
\bauthor{\bsnm{Wang}, \binits{Y.}},
\bauthor{\bsnm{Zhu}, \binits{C.}}:
\batitle{Peod: A pixel-aligned event-rgb benchmark for object detection under
  challenging conditions}.
\bjtitle{Proceedings of the AAAI Conference on Artificial Intelligence}
\bvolume{40}(\bissue{22}),
\bfpage{18207}--\blpage{18215}
(\byear{2026})
\end{barticle}
\endbibitem

\bibitem{wang2026evreid}
\begin{barticle}
\bauthor{\bsnm{Wang}, \binits{X.}},
\bauthor{\bsnm{Zhu}, \binits{Q.}},
\bauthor{\bsnm{Wu}, \binits{S.}},
\bauthor{\bsnm{Jiang}, \binits{B.}},
\bauthor{\bsnm{Zhang}, \binits{S.}}:
\batitle{When person re-identification meets event camera: A benchmark dataset
  and an attribute-guided re-identification framework}.
\bjtitle{Proceedings of the AAAI Conference on Artificial Intelligence}
\bvolume{40}(\bissue{12}),
\bfpage{10172}--\blpage{10180}
(\byear{2026})
\end{barticle}
\endbibitem

\bibitem{wang2024visevent}
\begin{barticle}
\bauthor{\bsnm{Wang}, \binits{X.}},
\bauthor{\bsnm{Li}, \binits{J.}},
\bauthor{\bsnm{Zhu}, \binits{L.}},
\bauthor{\bsnm{Zhang}, \binits{Z.}},
\bauthor{\bsnm{Chen}, \binits{Z.}},
\bauthor{\bsnm{Li}, \binits{X.}},
\bauthor{\bsnm{Wang}, \binits{Y.}},
\bauthor{\bsnm{Tian}, \binits{Y.}},
\bauthor{\bsnm{Wu}, \binits{F.}}:
\batitle{Visevent: Reliable object tracking via collaboration of frame and
  event flows}.
\bjtitle{IEEE Transactions on Cybernetics}
\bvolume{54}(\bissue{3}),
\bfpage{1997}--\blpage{2010}
(\byear{2024})
\end{barticle}
\endbibitem

\bibitem{zhu2025visible}
\begin{barticle}
\bauthor{\bsnm{Zhu}, \binits{Y.}},
\bauthor{\bsnm{Wang}, \binits{Q.}},
\bauthor{\bsnm{Li}, \binits{C.}},
\bauthor{\bsnm{Tang}, \binits{J.}},
\bauthor{\bsnm{Gu}, \binits{C.}},
\bauthor{\bsnm{Huang}, \binits{Z.}}:
\batitle{Visible--thermal multiple object tracking: Large-scale video dataset
  and progressive fusion approach}.
\bjtitle{Pattern Recognition}
\bvolume{161},
\bfpage{111330}
(\byear{2025})
\end{barticle}
\endbibitem

\bibitem{Zhang_2021_ICCV}
\begin{bchapter}
\bauthor{\bsnm{Zhang}, \binits{J.}},
\bauthor{\bsnm{Yang}, \binits{X.}},
\bauthor{\bsnm{Fu}, \binits{Y.}},
\bauthor{\bsnm{Wei}, \binits{X.}},
\bauthor{\bsnm{Yin}, \binits{B.}},
\bauthor{\bsnm{Dong}, \binits{B.}}:
\bctitle{Object tracking by jointly exploiting frame and event domain}.
In: \bbtitle{Proceedings of the IEEE/CVF International Conference on Computer
  Vision},
pp. \bfpage{13043}--\blpage{13052}
(\byear{2021})
\end{bchapter}
\endbibitem

\bibitem{tang2025revisiting}
\begin{botherref}
\oauthor{\bsnm{Tang}, \binits{C.}},
\oauthor{\bsnm{Wang}, \binits{X.}},
\oauthor{\bsnm{Huang}, \binits{J.}},
\oauthor{\bsnm{Jiang}, \binits{B.}},
\oauthor{\bsnm{Zhu}, \binits{L.}},
\oauthor{\bsnm{Chen}, \binits{S.}},
\oauthor{\bsnm{Zhang}, \binits{J.}},
\oauthor{\bsnm{Wang}, \binits{Y.}},
\oauthor{\bsnm{Tian}, \binits{Y.}}:
Revisiting color-event based tracking: A unified network, dataset, and metric.
Pattern Recognition,
112718
(2025)
\end{botherref}
\endbibitem

\bibitem{chen2024delving}
\begin{bchapter}
\bauthor{\bsnm{Chen}, \binits{S.}},
\bauthor{\bsnm{Yu}, \binits{E.}},
\bauthor{\bsnm{Li}, \binits{J.}},
\bauthor{\bsnm{Tao}, \binits{W.}}:
\bctitle{Delving into the trajectory long-tail distribution for multi-object
  tracking}.
In: \bbtitle{Proceedings of the IEEE/CVF Conference on Computer Vision and
  Pattern Recognition},
pp. \bfpage{19341}--\blpage{19351}
(\byear{2024})
\end{bchapter}
\endbibitem

\bibitem{kuhn1955hungarian}
\begin{barticle}
\bauthor{\bsnm{Kuhn}, \binits{H.W.}}:
\batitle{The hungarian method for the assignment problem}.
\bjtitle{Naval research logistics quarterly}
\bvolume{2}(\bissue{1-2}),
\bfpage{83}--\blpage{97}
(\byear{1955})
\end{barticle}
\endbibitem

\bibitem{qin2023motiontrack}
\begin{bchapter}
\bauthor{\bsnm{Qin}, \binits{Z.}},
\bauthor{\bsnm{Zhou}, \binits{S.}},
\bauthor{\bsnm{Wang}, \binits{L.}},
\bauthor{\bsnm{Duan}, \binits{J.}},
\bauthor{\bsnm{Hua}, \binits{G.}},
\bauthor{\bsnm{Tang}, \binits{W.}}:
\bctitle{Motiontrack: Learning robust short-term and long-term motions for
  multi-object tracking}.
In: \bbtitle{Proceedings of the IEEE/CVF Conference on Computer Vision and
  Pattern Recognition},
pp. \bfpage{17939}--\blpage{17948}
(\byear{2023})
\end{bchapter}
\endbibitem

\bibitem{han2022mat}
\begin{barticle}
\bauthor{\bsnm{Han}, \binits{S.}},
\bauthor{\bsnm{Huang}, \binits{P.}},
\bauthor{\bsnm{Wang}, \binits{H.}},
\bauthor{\bsnm{Yu}, \binits{E.}},
\bauthor{\bsnm{Liu}, \binits{D.}},
\bauthor{\bsnm{Pan}, \binits{X.}}:
\batitle{Mat: Motion-aware multi-object tracking}.
\bjtitle{Neurocomputing}
\bvolume{476},
\bfpage{75}--\blpage{86}
(\byear{2022})
\end{barticle}
\endbibitem

\bibitem{pang2021quasi}
\begin{bchapter}
\bauthor{\bsnm{Pang}, \binits{J.}},
\bauthor{\bsnm{Qiu}, \binits{L.}},
\bauthor{\bsnm{Li}, \binits{X.}},
\bauthor{\bsnm{Chen}, \binits{H.}},
\bauthor{\bsnm{Li}, \binits{Q.}},
\bauthor{\bsnm{Darrell}, \binits{T.}},
\bauthor{\bsnm{Yu}, \binits{F.}}:
\bctitle{Quasi-dense similarity learning for multiple object tracking}.
In: \bbtitle{Proceedings of the IEEE/CVF Conference on Computer Vision and
  Pattern Recognition},
pp. \bfpage{164}--\blpage{173}
(\byear{2021})
\end{bchapter}
\endbibitem

\bibitem{wang2021multiple}
\begin{bchapter}
\bauthor{\bsnm{Wang}, \binits{Q.}},
\bauthor{\bsnm{Zheng}, \binits{Y.}},
\bauthor{\bsnm{Pan}, \binits{P.}},
\bauthor{\bsnm{Xu}, \binits{Y.}}:
\bctitle{Multiple object tracking with correlation learning}.
In: \bbtitle{Proceedings of the IEEE/CVF Conference on Computer Vision and
  Pattern Recognition},
pp. \bfpage{3876}--\blpage{3886}
(\byear{2021})
\end{bchapter}
\endbibitem

\bibitem{wojke2017simple}
\begin{bchapter}
\bauthor{\bsnm{Wojke}, \binits{N.}},
\bauthor{\bsnm{Bewley}, \binits{A.}},
\bauthor{\bsnm{Paulus}, \binits{D.}}:
\bctitle{Simple online and realtime tracking with a deep association metric}.
In: \bbtitle{2017 IEEE International Conference on Image Processing (ICIP)},
pp. \bfpage{3645}--\blpage{3649}
(\byear{2017}).
\bcomment{IEEE}
\end{bchapter}
\endbibitem

\bibitem{aharon2022bot}
\begin{botherref}
\oauthor{\bsnm{Aharon}, \binits{N.}},
\oauthor{\bsnm{Orfaig}, \binits{R.}},
\oauthor{\bsnm{Bobrovsky}, \binits{B.-Z.}}:
Bot-sort: Robust associations multi-pedestrian tracking.
arXiv preprint arXiv:2206.14651
(2022)
\end{botherref}
\endbibitem

\bibitem{du2023strongsort}
\begin{barticle}
\bauthor{\bsnm{Du}, \binits{Y.}},
\bauthor{\bsnm{Zhao}, \binits{Z.}},
\bauthor{\bsnm{Song}, \binits{Y.}},
\bauthor{\bsnm{Zhao}, \binits{Y.}},
\bauthor{\bsnm{Su}, \binits{F.}},
\bauthor{\bsnm{Gong}, \binits{T.}},
\bauthor{\bsnm{Meng}, \binits{H.}}:
\batitle{Strongsort: Make deepsort great again}.
\bjtitle{IEEE Transactions on Multimedia}
\bvolume{25},
\bfpage{8725}--\blpage{8737}
(\byear{2023})
\end{barticle}
\endbibitem

\bibitem{qin2024towards}
\begin{bchapter}
\bauthor{\bsnm{Qin}, \binits{Z.}},
\bauthor{\bsnm{Wang}, \binits{L.}},
\bauthor{\bsnm{Zhou}, \binits{S.}},
\bauthor{\bsnm{Fu}, \binits{P.}},
\bauthor{\bsnm{Hua}, \binits{G.}},
\bauthor{\bsnm{Tang}, \binits{W.}}:
\bctitle{Towards generalizable multi-object tracking}.
In: \bbtitle{Proceedings of the IEEE/CVF Conference on Computer Vision and
  Pattern Recognition},
pp. \bfpage{18995}--\blpage{19004}
(\byear{2024})
\end{bchapter}
\endbibitem

\bibitem{bergmann2019tracking}
\begin{bchapter}
\bauthor{\bsnm{Bergmann}, \binits{P.}},
\bauthor{\bsnm{Meinhardt}, \binits{T.}},
\bauthor{\bsnm{Leal-Taixe}, \binits{L.}}:
\bctitle{Tracking without bells and whistles}.
In: \bbtitle{Proceedings of the IEEE/CVF International Conference on Computer
  Vision},
pp. \bfpage{941}--\blpage{951}
(\byear{2019})
\end{bchapter}
\endbibitem

\bibitem{liang2022rethinking}
\begin{barticle}
\bauthor{\bsnm{Liang}, \binits{C.}},
\bauthor{\bsnm{Zhang}, \binits{Z.}},
\bauthor{\bsnm{Zhou}, \binits{X.}},
\bauthor{\bsnm{Li}, \binits{B.}},
\bauthor{\bsnm{Zhu}, \binits{S.}},
\bauthor{\bsnm{Hu}, \binits{W.}}:
\batitle{Rethinking the competition between detection and reid in multiobject
  tracking}.
\bjtitle{IEEE Transactions on Image Processing}
\bvolume{31},
\bfpage{3182}--\blpage{3196}
(\byear{2022})
\end{barticle}
\endbibitem

\bibitem{wang2020towards}
\begin{bchapter}
\bauthor{\bsnm{Wang}, \binits{Z.}},
\bauthor{\bsnm{Zheng}, \binits{L.}},
\bauthor{\bsnm{Liu}, \binits{Y.}},
\bauthor{\bsnm{Li}, \binits{Y.}},
\bauthor{\bsnm{Wang}, \binits{S.}}:
\bctitle{Towards real-time multi-object tracking}.
In: \bbtitle{European Conference on Computer Vision},
pp. \bfpage{107}--\blpage{122}
(\byear{2020}).
\bcomment{Springer}
\end{bchapter}
\endbibitem

\bibitem{zhou2022global}
\begin{bchapter}
\bauthor{\bsnm{Zhou}, \binits{X.}},
\bauthor{\bsnm{Yin}, \binits{T.}},
\bauthor{\bsnm{Koltun}, \binits{V.}},
\bauthor{\bsnm{Kr{\"a}henb{\"u}hl}, \binits{P.}}:
\bctitle{Global tracking transformers}.
In: \bbtitle{Proceedings of the IEEE/CVF Conference on Computer Vision and
  Pattern Recognition},
pp. \bfpage{8771}--\blpage{8780}
(\byear{2022})
\end{bchapter}
\endbibitem

\bibitem{sun2020transtrack}
\begin{botherref}
\oauthor{\bsnm{Sun}, \binits{P.}},
\oauthor{\bsnm{Cao}, \binits{J.}},
\oauthor{\bsnm{Jiang}, \binits{Y.}},
\oauthor{\bsnm{Zhang}, \binits{R.}},
\oauthor{\bsnm{Xie}, \binits{E.}},
\oauthor{\bsnm{Yuan}, \binits{Z.}},
\oauthor{\bsnm{Wang}, \binits{C.}},
\oauthor{\bsnm{Luo}, \binits{P.}}:
Transtrack: Multiple object tracking with transformer.
arXiv preprint arXiv:2012.15460
(2020)
\end{botherref}
\endbibitem

\bibitem{zeng2022motr}
\begin{bchapter}
\bauthor{\bsnm{Zeng}, \binits{F.}},
\bauthor{\bsnm{Dong}, \binits{B.}},
\bauthor{\bsnm{Zhang}, \binits{Y.}},
\bauthor{\bsnm{Wang}, \binits{T.}},
\bauthor{\bsnm{Zhang}, \binits{X.}},
\bauthor{\bsnm{Wei}, \binits{Y.}}:
\bctitle{Motr: End-to-end multiple-object tracking with transformer}.
In: \bbtitle{European Conference on Computer Vision},
pp. \bfpage{659}--\blpage{675}
(\byear{2022}).
\bcomment{Springer}
\end{bchapter}
\endbibitem

\bibitem{meinhardt2022trackformer}
\begin{bchapter}
\bauthor{\bsnm{Meinhardt}, \binits{T.}},
\bauthor{\bsnm{Kirillov}, \binits{A.}},
\bauthor{\bsnm{Leal-Taixe}, \binits{L.}},
\bauthor{\bsnm{Feichtenhofer}, \binits{C.}}:
\bctitle{Trackformer: Multi-object tracking with transformers}.
In: \bbtitle{Proceedings of the IEEE/CVF Conference on Computer Vision and
  Pattern Recognition},
pp. \bfpage{8844}--\blpage{8854}
(\byear{2022})
\end{bchapter}
\endbibitem

\bibitem{leal2015motchallenge}
\begin{botherref}
\oauthor{\bsnm{Leal-Taix{\'e}}, \binits{L.}},
\oauthor{\bsnm{Milan}, \binits{A.}},
\oauthor{\bsnm{Reid}, \binits{I.}},
\oauthor{\bsnm{Roth}, \binits{S.}},
\oauthor{\bsnm{Schindler}, \binits{K.}}:
Motchallenge 2015: Towards a benchmark for multi-target tracking.
arXiv preprint arXiv:1504.01942
(2015)
\end{botherref}
\endbibitem

\bibitem{Gehrig21ral}
\begin{botherref}
\oauthor{\bsnm{Gehrig}, \binits{M.}},
\oauthor{\bsnm{Aarents}, \binits{W.}},
\oauthor{\bsnm{Gehrig}, \binits{D.}},
\oauthor{\bsnm{Scaramuzza}, \binits{D.}}:
Dsec: A stereo event camera dataset for driving scenarios.
IEEE Robotics and Automation Letters
(2021)
\end{botherref}
\endbibitem

\bibitem{tomy2022fusing}
\begin{bchapter}
\bauthor{\bsnm{Tomy}, \binits{A.}},
\bauthor{\bsnm{Paigwar}, \binits{A.}},
\bauthor{\bsnm{Mann}, \binits{K.S.}},
\bauthor{\bsnm{Renzaglia}, \binits{A.}},
\bauthor{\bsnm{Laugier}, \binits{C.}}:
\bctitle{Fusing event-based and rgb camera for robust object detection in
  adverse conditions}.
In: \bbtitle{2022 International Conference on Robotics and Automation (ICRA)},
pp. \bfpage{933}--\blpage{939}
(\byear{2022}).
\bcomment{IEEE}
\end{bchapter}
\endbibitem

\bibitem{zhou2023rgb}
\begin{bchapter}
\bauthor{\bsnm{Zhou}, \binits{Z.}},
\bauthor{\bsnm{Wu}, \binits{Z.}},
\bauthor{\bsnm{Boutteau}, \binits{R.}},
\bauthor{\bsnm{Yang}, \binits{F.}},
\bauthor{\bsnm{Demonceaux}, \binits{C.}},
\bauthor{\bsnm{Ginhac}, \binits{D.}}:
\bctitle{Rgb-event fusion for moving object detection in autonomous driving}.
In: \bbtitle{2023 IEEE International Conference on Robotics and Automation
  (ICRA)},
pp. \bfpage{7808}--\blpage{7815}
(\byear{2023}).
\bcomment{IEEE}
\end{bchapter}
\endbibitem

\bibitem{zhou2024bring}
\begin{bchapter}
\bauthor{\bsnm{Zhou}, \binits{H.}},
\bauthor{\bsnm{Chang}, \binits{Y.}},
\bauthor{\bsnm{Shi}, \binits{Z.}}:
\bctitle{Bring event into rgb and lidar: Hierarchical visual-motion fusion for
  scene flow}.
In: \bbtitle{Proceedings of the IEEE/CVF Conference on Computer Vision and
  Pattern Recognition},
pp. \bfpage{26477}--\blpage{26486}
(\byear{2024})
\end{bchapter}
\endbibitem

\bibitem{gehrig2024low}
\begin{barticle}
\bauthor{\bsnm{Gehrig}, \binits{D.}},
\bauthor{\bsnm{Scaramuzza}, \binits{D.}}:
\batitle{Low-latency automotive vision with event cameras}.
\bjtitle{Nature}
\bvolume{629}(\bissue{8014}),
\bfpage{1034}--\blpage{1040}
(\byear{2024})
\end{barticle}
\endbibitem

\bibitem{liu2025beyond}
\begin{barticle}
\bauthor{\bsnm{Liu}, \binits{Z.}},
\bauthor{\bsnm{Sun}, \binits{Y.}},
\bauthor{\bsnm{Wang}, \binits{Y.}},
\bauthor{\bsnm{Yang}, \binits{N.}},
\bauthor{\bsnm{Li}, \binits{S.E.}},
\bauthor{\bsnm{Zhao}, \binits{X.}}:
\batitle{Beyond conventional vision: Rgb-event fusion for robust object
  detection in dynamic traffic scenarios}.
\bjtitle{Communications in Transportation Research}
\bvolume{5},
\bfpage{100202}
(\byear{2025})
\end{barticle}
\endbibitem

\bibitem{he2016deep}
\begin{bchapter}
\bauthor{\bsnm{He}, \binits{K.}},
\bauthor{\bsnm{Zhang}, \binits{X.}},
\bauthor{\bsnm{Ren}, \binits{S.}},
\bauthor{\bsnm{Sun}, \binits{J.}}:
\bctitle{Deep residual learning for image recognition}.
In: \bbtitle{Proceedings of the IEEE Conference on Computer Vision and Pattern
  Recognition},
pp. \bfpage{770}--\blpage{778}
(\byear{2016})
\end{bchapter}
\endbibitem

\bibitem{X-AnyLabeling}
\begin{botherref}
\oauthor{\bsnm{Wang}, \binits{W.}}:
Advanced Auto Labeling Solution with Added Features.
Github
(2023)
\end{botherref}
\endbibitem

\bibitem{van2008visualizing}
\begin{botherref}
\oauthor{\bparticle{Van~der} \bsnm{Maaten}, \binits{L.}},
\oauthor{\bsnm{Hinton}, \binits{G.}}:
Visualizing data using t-sne.
Journal of machine learning research
\textbf{9}(11)
(2008)
\end{botherref}
\endbibitem

\bibitem{cao2023observation}
\begin{bchapter}
\bauthor{\bsnm{Cao}, \binits{J.}},
\bauthor{\bsnm{Pang}, \binits{J.}},
\bauthor{\bsnm{Weng}, \binits{X.}},
\bauthor{\bsnm{Khirodkar}, \binits{R.}},
\bauthor{\bsnm{Kitani}, \binits{K.}}:
\bctitle{Observation-centric sort: Rethinking sort for robust multi-object
  tracking}.
In: \bbtitle{Proceedings of the IEEE/CVF Conference on Computer Vision and
  Pattern Recognition},
pp. \bfpage{9686}--\blpage{9696}
(\byear{2023})
\end{bchapter}
\endbibitem

\bibitem{yang2024hybrid}
\begin{bchapter}
\bauthor{\bsnm{Yang}, \binits{M.}},
\bauthor{\bsnm{Han}, \binits{G.}},
\bauthor{\bsnm{Yan}, \binits{B.}},
\bauthor{\bsnm{Zhang}, \binits{W.}},
\bauthor{\bsnm{Qi}, \binits{J.}},
\bauthor{\bsnm{Lu}, \binits{H.}},
\bauthor{\bsnm{Wang}, \binits{D.}}:
\bctitle{Hybrid-sort: Weak cues matter for online multi-object tracking}.
In: \bbtitle{Proceedings of the AAAI Conference on Artificial Intelligence},
vol. \bseriesno{38},
pp. \bfpage{6504}--\blpage{6512}
(\byear{2024})
\end{bchapter}
\endbibitem

\bibitem{li2024sampling}
\begin{bchapter}
\bauthor{\bsnm{Li}, \binits{Z.}},
\bauthor{\bsnm{Zhang}, \binits{D.}},
\bauthor{\bsnm{Wu}, \binits{S.}},
\bauthor{\bsnm{Song}, \binits{M.}},
\bauthor{\bsnm{Chen}, \binits{G.}}:
\bctitle{Sampling-resilient multi-object tracking}.
In: \bbtitle{Proceedings of the AAAI Conference on Artificial Intelligence},
vol. \bseriesno{38},
pp. \bfpage{3297}--\blpage{3305}
(\byear{2024})
\end{bchapter}
\endbibitem

\bibitem{wu2021track}
\begin{bchapter}
\bauthor{\bsnm{Wu}, \binits{J.}},
\bauthor{\bsnm{Cao}, \binits{J.}},
\bauthor{\bsnm{Song}, \binits{L.}},
\bauthor{\bsnm{Wang}, \binits{Y.}},
\bauthor{\bsnm{Yang}, \binits{M.}},
\bauthor{\bsnm{Yuan}, \binits{J.}}:
\bctitle{Track to detect and segment: An online multi-object tracker}.
In: \bbtitle{Proceedings of the IEEE/CVF Conference on Computer Vision and
  Pattern Recognition},
pp. \bfpage{12352}--\blpage{12361}
(\byear{2021})
\end{bchapter}
\endbibitem

\bibitem{zhang2023motrv2}
\begin{bchapter}
\bauthor{\bsnm{Zhang}, \binits{Y.}},
\bauthor{\bsnm{Wang}, \binits{T.}},
\bauthor{\bsnm{Zhang}, \binits{X.}}:
\bctitle{Motrv2: Bootstrapping end-to-end multi-object tracking by pretrained
  object detectors}.
In: \bbtitle{Proceedings of the IEEE/CVF Conference on Computer Vision and
  Pattern Recognition},
pp. \bfpage{22056}--\blpage{22065}
(\byear{2023})
\end{bchapter}
\endbibitem

\bibitem{bernardin2008evaluating}
\begin{barticle}
\bauthor{\bsnm{Bernardin}, \binits{K.}},
\bauthor{\bsnm{Stiefelhagen}, \binits{R.}}:
\batitle{Evaluating multiple object tracking performance: the clear mot
  metrics}.
\bjtitle{EURASIP Journal on Image and Video Processing}
\bvolume{2008},
\bfpage{1}--\blpage{10}
(\byear{2008})
\end{barticle}
\endbibitem

\bibitem{luiten2021hota}
\begin{barticle}
\bauthor{\bsnm{Luiten}, \binits{J.}},
\bauthor{\bsnm{Osep}, \binits{A.}},
\bauthor{\bsnm{Dendorfer}, \binits{P.}},
\bauthor{\bsnm{Torr}, \binits{P.}},
\bauthor{\bsnm{Geiger}, \binits{A.}},
\bauthor{\bsnm{Leal-Taix{\'e}}, \binits{L.}},
\bauthor{\bsnm{Leibe}, \binits{B.}}:
\batitle{Hota: A higher order metric for evaluating multi-object tracking}.
\bjtitle{International journal of computer vision}
\bvolume{129},
\bfpage{548}--\blpage{578}
(\byear{2021})
\end{barticle}
\endbibitem

\bibitem{ristani2016performance}
\begin{bchapter}
\bauthor{\bsnm{Ristani}, \binits{E.}},
\bauthor{\bsnm{Solera}, \binits{F.}},
\bauthor{\bsnm{Zou}, \binits{R.}},
\bauthor{\bsnm{Cucchiara}, \binits{R.}},
\bauthor{\bsnm{Tomasi}, \binits{C.}}:
\bctitle{Performance measures and a data set for multi-target, multi-camera
  tracking}.
In: \bbtitle{European Conference on Computer Vision},
pp. \bfpage{17}--\blpage{35}
(\byear{2016}).
\bcomment{Springer}
\end{bchapter}
\endbibitem

\bibitem{liu2022dabdetr}
\begin{bchapter}
\bauthor{\bsnm{Liu}, \binits{S.}},
\bauthor{\bsnm{Li}, \binits{F.}},
\bauthor{\bsnm{Zhang}, \binits{H.}},
\bauthor{\bsnm{Yang}, \binits{X.}},
\bauthor{\bsnm{Qi}, \binits{X.}},
\bauthor{\bsnm{Su}, \binits{H.}},
\bauthor{\bsnm{Zhu}, \binits{J.}},
\bauthor{\bsnm{Zhang}, \binits{L.}}:
\bctitle{{DAB}-{DETR}: Dynamic anchor boxes are better queries for {DETR}}.
In: \bbtitle{International Conference on Learning Representations}
(\byear{2022})
\end{bchapter}
\endbibitem

\bibitem{loshchilov2018adamw}
\begin{bchapter}
\bauthor{\bsnm{Loshchilov}, \binits{I.}},
\bauthor{\bsnm{Hutter}, \binits{F.}}:
\bctitle{Decoupled weight decay regularization}.
In: \bbtitle{International Conference on Learning Representations}
(\byear{2019})
\end{bchapter}
\endbibitem

\bibitem{paszke2019pytorch}
\begin{botherref}
\oauthor{\bsnm{Paszke}, \binits{A.}},
\oauthor{\bsnm{Gross}, \binits{S.}},
\oauthor{\bsnm{Massa}, \binits{F.}},
\oauthor{\bsnm{Lerer}, \binits{A.}},
\oauthor{\bsnm{Bradbury}, \binits{J.}},
\oauthor{\bsnm{Chanan}, \binits{G.}},
\oauthor{\bsnm{Killeen}, \binits{T.}},
\oauthor{\bsnm{Lin}, \binits{Z.}},
\oauthor{\bsnm{Gimelshein}, \binits{N.}},
\oauthor{\bsnm{Antiga}, \binits{L.}}, et al.:
Pytorch: An imperative style, high-performance deep learning library.
Advances in neural information processing systems
\textbf{32}
(2019)
\end{botherref}
\endbibitem

\bibitem{shuai2021siammot}
\begin{bchapter}
\bauthor{\bsnm{Shuai}, \binits{B.}},
\bauthor{\bsnm{Berneshawi}, \binits{A.}},
\bauthor{\bsnm{Li}, \binits{X.}},
\bauthor{\bsnm{Modolo}, \binits{D.}},
\bauthor{\bsnm{Tighe}, \binits{J.}}:
\bctitle{Siammot: Siamese multi-object tracking}.
In: \bbtitle{Proceedings of the IEEE/CVF Conference on Computer Vision and
  Pattern Recognition},
pp. \bfpage{12372}--\blpage{12382}
(\byear{2021})
\end{bchapter}
\endbibitem

\bibitem{ren2024frequency}
\begin{botherref}
\oauthor{\bsnm{Ren}, \binits{H.}},
\oauthor{\bsnm{Ma}, \binits{F.}},
\oauthor{\bsnm{Lin}, \binits{X.}},
\oauthor{\bsnm{Fang}, \binits{Y.}},
\oauthor{\bsnm{Huang}, \binits{H.}},
\oauthor{\bsnm{Huang}, \binits{Y.}},
\oauthor{\bsnm{Zhou}, \binits{Y.}},
\oauthor{\bsnm{Fu}, \binits{H.}},
\oauthor{\bsnm{Yang}, \binits{Z.}},
\oauthor{\bsnm{Yu}, \binits{F.R.}}, et al.:
Frequency-aware event cloud network.
arXiv preprint arXiv:2412.20803
(2024)
\end{botherref}
\endbibitem

\bibitem{kim2024frequency}
\begin{bchapter}
\bauthor{\bsnm{Kim}, \binits{T.}},
\bauthor{\bsnm{Cho}, \binits{H.}},
\bauthor{\bsnm{Yoon}, \binits{K.-J.}}:
\bctitle{Frequency-aware event-based video deblurring for real-world motion
  blur}.
In: \bbtitle{Proceedings of the IEEE/CVF Conference on Computer Vision and
  Pattern Recognition},
pp. \bfpage{24966}--\blpage{24976}
(\byear{2024})
\end{bchapter}
\endbibitem

\bibitem{xu2025learning}
\begin{botherref}
\oauthor{\bsnm{Xu}, \binits{B.}},
\oauthor{\bsnm{Hou}, \binits{R.}},
\oauthor{\bsnm{Ren}, \binits{T.}},
\oauthor{\bsnm{Wu}, \binits{G.}},
\oauthor{\bsnm{Cao}, \binits{J.}}, et al.:
Learning frequency and memory-aware prompts for multi-modal object tracking.
arXiv preprint arXiv:2506.23972
(2025)
\end{botherref}
\endbibitem

\end{thebibliography}


\end{document}